# Co-Occurrence Priors Override Visual Spatial Evidence: Understanding False Alarms in Vision–Language Models for Construction Struck-By Hazard Detection

**Xiaoshan Zhou**

School of Project Management, Faculty of Engineering, University of Sydney, NSW, Australia, 2037, xiaoshan.zhou@sydney.edu.au

## Abstract

Vision–language models (VLMs) increasingly support construction hazard monitoring, yet their frequent false alarms remain poorly understood. This study investigates whether false struck-by hazard warnings arise from failures in object recognition, spatial relation encoding, or hazard decisions dominated by learned co-occurrence priors. We evaluated Qwen2.5-VL-7B and InternVL3.5-8B using 355 counterfactual worker–excavator scene pairs with controlled spatial proximity and worker-removal variants, combining behavioral analysis, internal representation probing, causal interventions, and decision correction. Both models recognized the entities and encoded their spatial relations (out-of-fold AUROC: 0.82 and 0.91), assigning higher hazard scores to near than far scenes in 79% and 73% of pairs. Nevertheless, 74% and 100% of far scenes were falsely classified as hazardous. Removing the worker reduced Qwen's hazard rate to 11%, while worker presence exerted 2.4- and 3.2-times greater influence on hazard scores than spatial proximity in the respective models. Attention-head interventions, neuron-level interventions, and activation steering failed to eliminate this co-occurrence bias. However, conditioning hazard decisions on the model's own spatial-relation representation improved Qwen's specificity from 0.36 to 0.58 at matched sensitivity, outperforming threshold recalibration. These findings reveal that VLM struck-by false alarms stem primarily from decision-making failures rather than deficiencies in object recognition or spatial encoding, suggesting that future efforts should prioritize conditioning hazard decisions on encoded spatial evidence rather than further improving visual recognition alone.

## 1. Introduction

Struck-by incidents are one of the four leading causes of construction fatalities identified by the US Occupational Safety and Health Administration (OSHA, n.d.). In 2024, 1,034 workers died in the US construction industry, more than in any other private industry, and transportation and contact incidents, which include being struck by vehicles, mobile equipment and objects, accounted for 39% of these deaths (BLS, 2026). Excavators are a characteristic source of this risk. Their booms, buckets and rotating superstructures sweep a large envelope around the machine, operators have restricted sight lines, and ground workers often enter the envelope to guide, measure or clean (Hinze and Teizer, 2011). Identifying worker–excavator struck-by hazards has therefore long been a central topic in construction safety research, particularly using computer vision. Detectors and segmentation models identify workers, protective equipment and plant in site imagery (Fang et al., 2018; Son et al., 2019), and proximity or relation modules estimate worker–equipment distances or interactions from fixed cameras, unmanned aerial vehicles or video (Kim et al., 2016; Kim et al., 2019; Tang et al., 2020; Xiong et al., 2019).

More recently, vision–language models (VLMs) have extended construction hazard recognition with end-to-end pipelines that can compare visual evidence with safety rules and directly answer natural-language questions, such as whether the visible worker is inside the excavator's swing zone. Construction researchers have begun to use VLMs and multimodal large language models for hazard querying, inspection and safety monitoring (Chen et al., 2024; Chan et al., 2025; Kim et al., 2025; Tsai et al., 2025; Wang et al., 2025c). However, current results reveal clear limitations: off-the-shelf models judged site safety at close to chance level before fine-tuning (Kim et al., 2025), produced convincing reasoning while identifying hazards poorly (Wang et al., 2025c), and flagged more safety violations than experts did, with high recall but low precision (Chen and Zou, 2026).

These problems may share a common root cause revealed by VLM research outside construction, which shows that VLMs often answer from language and co-occurrence statistics rather than from the image. Visual question-answering models exploit answer priors (Goyal et al., 2017; Agrawal et al., 2018), and captioning and instruction-tuned models report objects that typically co-occur with what is shown but are absent (Rohrbach et al., 2018; Li et al., 2023b; Zhou et al., 2024). In addition, VLMs perform poorly on spatial relations such as left/right, above/below or near/far, and sometimes behave like bags of words that register which entities are mentioned but not how they are arranged (Kamath et al., 2023; Liu et al., 2023; Yuksekgonul et al., 2023; Tong et al., 2024). These vulnerabilities complicate the diagnosis of VLM failures in struck-by hazard recognition. A valid hazard requires two entities, a worker and an excavator, to co-occur in the image, and the model must recognize the spatial relation between the worker and the excavator's operating zone before routing that evidence to its language-based decision. No study has yet established what causes VLM failures in hazard recognition, and the direction of remedy therefore remains unclear.

More specifically, when a VLM wrongly labels a worker standing well clear of an excavator as being in danger, at least three explanations are possible, and they call for different remedies. The model may fail to see the worker or the machine (a perception failure, addressed by better visual encoders or higher resolution). It may see both but fail to encode where the worker is relative to the machine (a relation failure, addressed by spatially grounded training data or geometric modules). Or it may see both and encode their relation, yet answer from a learned association between worker–excavator co-occurrence and danger that overrides the encoded relation (a decision failure, addressed by recalibration or by routing relational evidence into the decision). Benchmark accuracy, the dominant evaluation criterion in current research, cannot separate these explanations, because all three produce the same wrong answer.

This study performs that diagnosis. To the best of our knowledge, it is the first study in construction to examine the decision-making behavior of VLMs through mechanistic interpretability. Evidence comes from two open-weight VLM families, Qwen2.5-VL-7B-Instruct and InternVL3.5-8B, and from a construction image dataset that we constructed to hold the scene, the machine and the worker's appearance fixed while varying the worker's placement. The entities and their spatial relation were decoded from the models' internal activations with pair-grouped probes. We further tested which prompt words, attention heads and residual states the decision depends on, evaluated correction strategies, including causal interventions, against threshold recalibration, and summarized the results with two interpretable quantities: the prior dominance index and the ratio of co-occurrence to relation increments. The results show that struck-by false alarms arise from a co-occurrence offset that dominates an encoded spatial relation, rather than from missed objects or a misread relation.

The remainder of the paper is organized as follows. Section 2 reviews machine-vision approaches to struck-by hazards, evidence on visual grounding versus language and co-occurrence priors in VLMs, and interpretability methods for examining what VLMs encode and use. Section 3 states the hypotheses and research questions and describes the models, the construction of the counterfactual and control datasets, and the behavioral, representation, causal and correction analyses. Section 4 reports the results in the order of the diagnostic framework, from behavior to internal representation, co-occurrence controls, natural-image transfer, causal localization and correction. Section 5 interprets the findings, examines whether the correction analyses resolve the problem and discusses implications for construction safety practice. Section 6 concludes with the main findings, recommendations, limitations and directions for future research.

# 2. Literature review

## 2.1 Machine vision for struck-by hazards

The development of computer vision for construction safety has followed the same sequence in which recognition failures are diagnosed in this study: from perceiving objects to reasoning about their relations. Early systems focused on object recognition and segmentation (Fang et al., 2020; Zhong et al., 2019). Deep detectors locate workers, protective equipment and plant in images and video under varying pose, scale and background (Fang et al., 2018; Son et al., 2019). For struck-by hazards in particular, spatial relation reasoning was later added on top of detection. Kim et al. (2016) combined object trajectories with fuzzy inference to assess the danger posed by moving equipment. Tang et al. (2020) argued that bounding-box detection followed by hand-crafted rules limits vision-based safety inspection, and proposed a human–object interaction model that classifies how workers interact with tools and equipment. Xiong et al. (2019) detected visual relationships between workers and site components in video, encoded both the observed operations and safety-guideline semantics as subject–predicate–object triples supported by a construction-safety ontology, and reasoned over them with logic rules to identify hazards such as failing to wear a helmet and walking beneath a crane.

These machine-vision systems share a structure that VLMs promise to replace. They detect each entity, estimate a geometric quantity such as distance or overlap, and apply a manually specified rule, such as a proximity threshold. The reasoning component is narrow: each rule must be designed, each detector trained on annotated site data, and each threshold calibrated for camera geometry. Adding a language component potentially strengthens this reasoning, allowing decisions in more complex situations to be expressed and justified in natural language. Since their release, VLMs have been introduced in construction safety for safety querying in augmented reality (Chen et al., 2024), zero-shot and caption-based inspection of protective equipment and site conditions (Gil and Lee, 2024; Tsai et al., 2025), ontology-guided agents for monitoring work at height (Chan et al., 2025), domain-knowledge-augmented inspection (Wang et al., 2025c), retrieval-augmented hazard detection (Orooje et al., 2026) and fine-tuning for context-aware safety assessment (Kim et al., 2025).

These studies confirm the value of VLMs, but their accuracy has fallen short of expectations. Off-the-shelf GPT-4V and LLaVA-1.5 judged whether scenes were safe with 53% and 48% accuracy, close to chance, before fine-tuning (Kim et al., 2025). A multimodal inspection system reasoned faithfully to safety rules yet identified hazards correctly in only 57% of cases (Wang et al., 2025c). Unaided VLMs fabricated hazards until retrieval and checklist verification were added (Orooje et al., 2026). On the ConstructionSite-10k benchmark, general VLMs flagged more rule violations than experts did, showing high recall but low precision (Chen and Zou, 2026). Researchers have improved accuracy through better prompting, domain ontologies, retrieval and fine-tuning. More fundamentally, however, the community has not established how language and vision work together in these models: whether the language component is used for rule retrieval, as the driver of reasoning, for producing natural-language output, or in some other role. This uncertainty also confounds the diagnosis of failures, which could arise from visual perception, from language reasoning or from a conflict between the two modalities. The present study aims to shed light on these questions.

## 2.2 Visual evidence versus language and co-occurrence priors in VLMs

Substantial evidence shows that multimodal models can answer without looking, which suggests that VLMs used for construction safety may suffer from similar problems. Early work in visual question answering showed that models exploit the statistical association between question types and answers: balanced datasets that pair each question with images giving different answers (Goyal et al., 2017) and test splits with shifted answer priors (Agrawal et al., 2018) exposed this reliance. Image captioning models hallucinate objects that typically co-occur with the depicted scene (Rohrbach et al., 2018), and object co-occurrence statistics in training data are a principal driver of such hallucination in both captioners (Biten et al., 2022) and instruction-tuned large VLMs (Li et al., 2023b; Zhou et al., 2024). Benchmarks designed to separate language-driven from vision-driven answers find that VLMs follow textual expectation when the image contradicts it (Guan et al., 2024), and decoding strategies that contrast predictions made with and without reliable visual input reduce, but do not eliminate, these errors (Leng et al., 2024).

Recent studies have made this problem more precise and shown that it persists in current models. Using counterfactual images that separate language priors from perception and commonsense, VLind-Bench found that almost all large VLMs relied heavily on language priors (Lee et al., 2025). When visual and textual evidence conflicted in vision-centric tasks, ten VLMs trusted the text far more than the image (Deng et al., 2025). Object co-occurrence has likewise been formalised as a spurious causal correlation: counterfactual benchmarks show that all mainstream large VLMs are susceptible to it (Xu et al., 2025), and hallucinated objects correspond to clusters of image tokens that frequently co-occur in training data (Wang et al., 2025a). Moreover, recent literature shows that visual evidence is often present inside the model but not used. VLMs perform substantially worse on vision-centric tasks than a direct readout of their own visual encoders, because the language model does not exploit visual information that is accessible throughout the network (Fu et al., 2025). Deep layers attend reliably to the correct evidence region even when the final answer is wrong (Liu et al., 2026b), attention falls on the correct location of small objects that the model nevertheless misidentifies (Zhang et al., 2025a), and visually grounded tokens that are never generated remain highly ranked in intermediate layers (Li et al., 2025). These findings describe a dissociation between what a model encodes and what it reports. However, whether this dissociation, observed for object presence and attributes in everyday images, also governs construction hazard recognition remains unknown.

In addition, reading spatial relations has been identified as a particular weakness in VLMs. VLMs perform near chance on many relational prepositions (Kamath et al., 2023; Liu et al., 2023), and often behave like bags of words that encode which entities a caption mentions but not their order or relation (Yuksekgonul et al., 2023). Multimodal large language models inherit visual shortcomings from their encoders (Tong et al., 2024) and can fail simple geometric judgments, such as whether two shapes overlap (Rahmanzadehgervi et al., 2024). Recent benchmarks confirm that spatial reasoning remains a bottleneck for video-based spatial question answering (Yang et al., 2025) and for three-dimensional reasoning about height, orientation and location (Ma et al., 2025). Although these weaknesses are known in general terms, it is not known how much of the failure of VLMs to recognize struck-by hazards can be attributed to weak spatial reasoning and how much to co-occurrence priors. No empirical study has shown whether VLMs can recognize that a worker is within the swing radius of an excavator, how this relation is encoded in their internal representations, or how it is routed to the downstream decision of whether a scene is hazardous; this study addresses these questions.

## 2.3 Interpreting what VLMs encode and use

VLMs are often treated as black boxes, but interpretability methods now provide a lens on their internal representations and decision-making behavior. In early work, linear probes trained on intermediate activations tested whether a property is linearly decodable at a given layer (Alain and Bengio, 2017). Later studies pointed out that decodability does not imply use, because a probe may exploit correlated features, so control tasks and careful validation are needed (Hewitt and Liang, 2019; Belinkov, 2022). Causal methods were introduced to address this limitation. Activation patching and causal tracing replace internal states from one input with those from a contrasting input and measure the effect on the output (Meng et al., 2022; Vig et al., 2020), and circuit analyses identify attention heads and pathways that implement a behavior (Wang et al., 2023). Sparse autoencoders decompose activations into many sparsely active features that are often more interpretable than individual neurons (Bricken et al., 2023; Cunningham et al., 2023; Gao et al., 2025), and steering adds or removes directions in activation space to change behavior at inference time (Li et al., 2023a; Zou et al., 2023).

These methods have recently been adapted to VLMs, and the adaptations motivate the analyses used in this study. Causal tracing, which exchanges internal states between inputs, has been transferred from language models to the BLIP VLM (Palit et al., 2023) and to multimodal large language models, where it showed that a small subset of visual tokens transfers image information to the layers that answer a question (Basu et al., 2024). Token ablation and attention knockout showed that LLaVA reads object information from object-specific visual tokens, whose removal reduced object identification accuracy by more than 70%, and assembles it at the final token position (Neo et al., 2025). Similar knockout analyses showed that image and question information are integrated in the lower and middle layers before being carried to the final position (Zhang et al., 2025b), with cross-modal information flow concentrated in roughly the middle quarter of the layers (Kaduri et al., 2025). Sparse autoencoders trained on vision–language representations yield features that are more monosemantic than individual neurons (Pach et al., 2025), and steering the residual stream along visual-information directions has reduced hallucination in VLMs (Li et al., 2025). These studies show that probes, token ablation, head knockout, residual-stream patching, sparse autoencoders and steering can locate where visual information is held in a VLM and where it influences the answer. They have, however, been applied to object identification, factual recall and object hallucination, in which the relevant evidence is the presence of an object. Whether the same tools can separate the encoding of a spatial relation from its use in a safety decision has not been examined.

Interpretability has not, to our knowledge, been applied to diagnose VLM failures in construction safety, although it promises to clarify VLM behavior in the way that neuroscience studies have clarified human decision-making. Such a diagnosis requires controlled experimental conditions that isolate the factor of interest; otherwise, a probe or patching effect may reflect differences between scenes rather than the spatial relation. It also requires validation that respects the paired structure of the data, so that a relation readout cannot succeed by recognizing scenes. The following sections describe how this study meets these requirements and what the resulting interpretability analyses reveal.

# 3. Methodology

## 3.1 Research framework and hypotheses

A VLM that answers a struck-by query must complete three operations in sequence: it must perceive the worker and the excavator, encode the spatial relation between them, and use that relation when selecting the answer (Fig. 1). A false alarm on an image in which the worker stands clear of the machine can therefore originate at any of three stages, and each origin implies a distinct observable signature. Under a perception failure (H1), the representations of the worker or the excavator are weak; because a missed worker removes the reason for an alarm, H1 also predicts that errors are dominated by missed hazards rather than false alarms. Under a relation failure (H2), the entities are represented but the near/far relation cannot be recovered from the model's internal states, and the answer does not vary systematically when the worker is moved. Under a co-occurrence override (H3), both the entities and the relation are encoded and the answer score moves in the correct direction when the worker is moved, yet images with a distant worker are still classified as hazardous because the score is dominated by a component that depends on the joint presence of the two entities rather than on their arrangement.

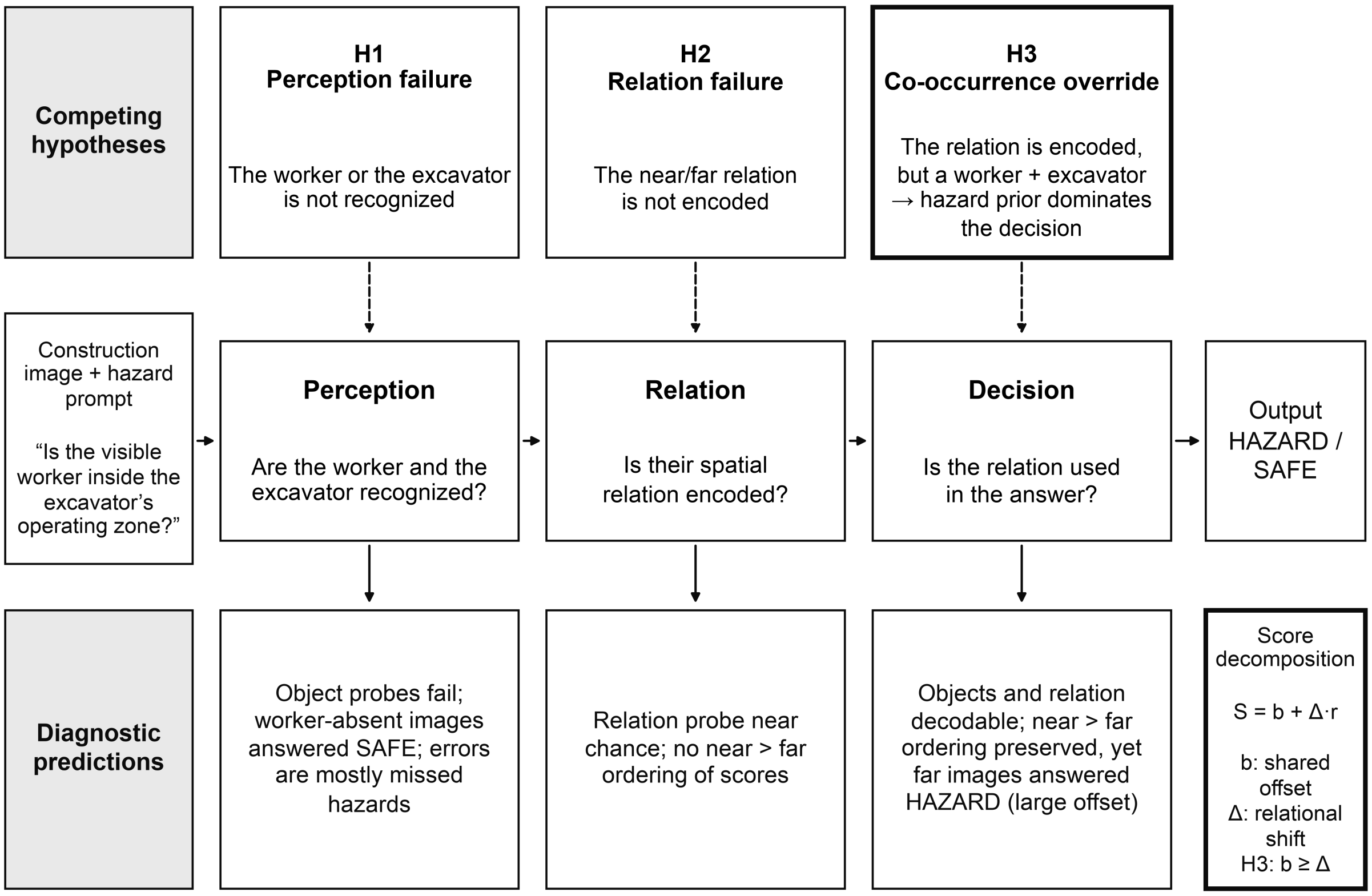

**Fig. 1.** Competing explanations for VLM struck-by errors and their diagnostic predictions. Perception failure (H1), relation failure (H2) and co-occurrence override (H3) predict different patterns of internal decodability and behavior. Under H3 the decision score $S$ decomposes into a shared offset $b$ and a relational shift $\Delta$, and errors on far images arise when $b$ is comparable to or larger than $\Delta$.

To express H3 quantitatively, let $x$ denote an image–prompt input and let $S(x)$ be the model's decision score, defined as the log-odds of HAZARD over SAFE (Section 3.2). The binary answer is

$$\hat{y}(x) = \mathbb{1}[S(x) > \tau], \quad \tau = 0 \tag{1}$$

where $\mathbb{1}[\cdot]$ is the indicator function and $\tau$ is the decision threshold implied by the model's own answer. For a scene presented with the same worker at a far ($r = 0$) or near ($r = 1$) placement, the score is decomposed as

$$S(x) = b + \Delta r(x) + \varepsilon(x) \tag{2}$$

where $b$ is the expected score of far images, Δ is the expected within-scene change produced by moving the worker into the operating zone, and $\varepsilon$ is a zero-mean residual. Under Eq. (2), a far image is classified correctly only if $b + \varepsilon < 0$. A positive offset $b$ thus produces false alarms that the relational term cannot remove, however reliably Δ is encoded. The relative size of the two terms is summarized by the prior dominance index (PDI),

$$\text{PDI} = \frac{b}{\Delta} \tag{3}$$

which expresses the offset in units of the relational shift. A PDI near or above one indicates that the relation-independent component is at least as influential as the relation itself. Section 3.4.3 further partitions $b$ into the contribution of the scene and prompt and the increment produced by adding a worker.

An overview of the study design is shown in Fig. 2. Its components jointly address four research questions:

- RQ1. Do the VLMs recognize the worker and the excavator and encode their near/far relation?
- RQ2. Do their HAZARD/SAFE answers use that relation, and how large is the relation-independent offset?
- RQ3. Is the offset produced by the co-occurrence of the two entities, by the scene and prompt alone, or by the specific relation?
- RQ4. Can internal relation evidence be used to correct the decision, and do model-internal components that sustain false alarms exist?

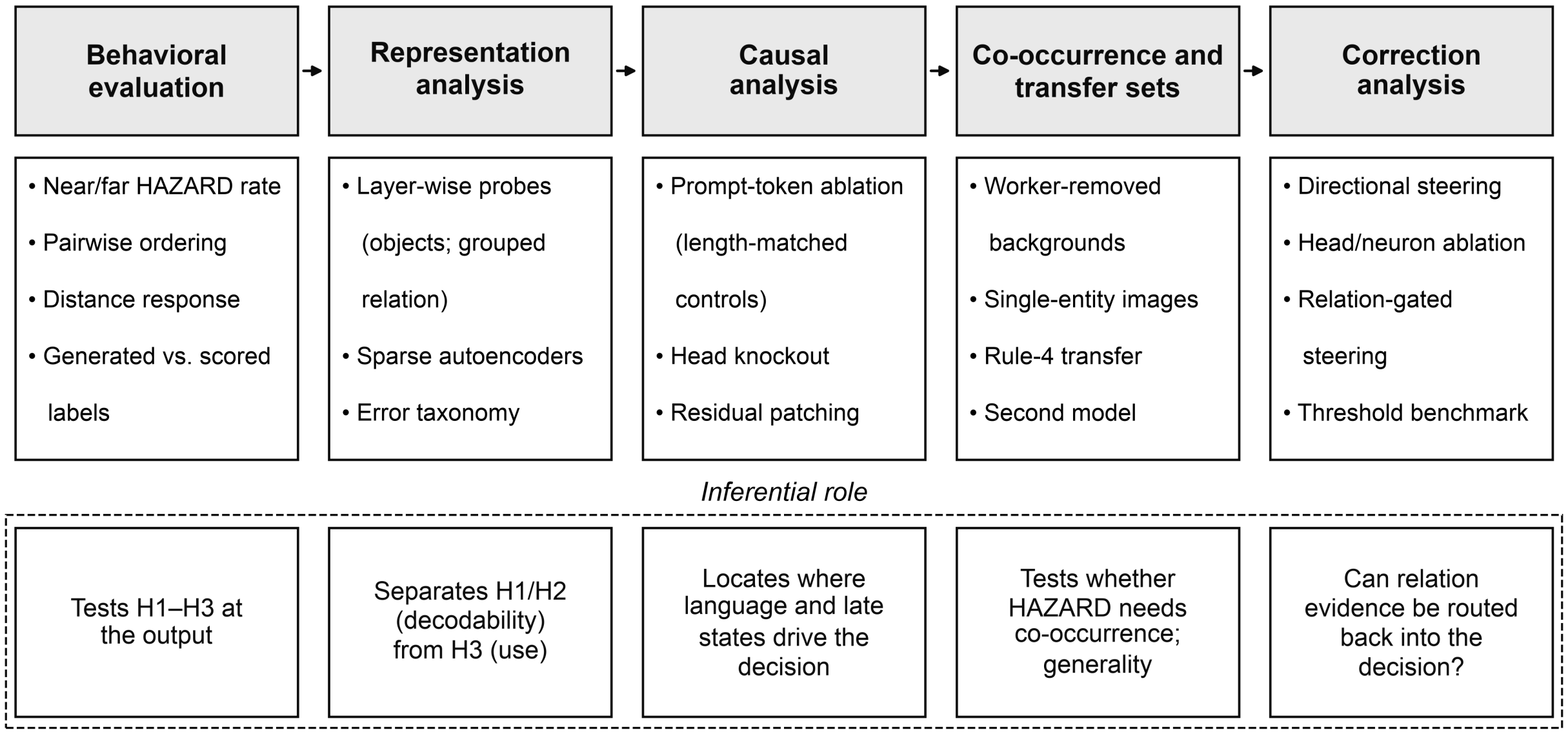

**Fig. 2.** Analysis workflow.

## 3.2 Models, prompt and decision scores

Two open-weight, instruction-tuned VLMs of comparable size but different architecture were evaluated: Qwen2.5-VL-7B-Instruct (Bai et al., 2025) and InternVL3.5-8B (Wang et al., 2025b). Both represent current open-weight families that couple a vision encoder, a multimodal projector and a decoder-only language model. All inference was performed in bfloat16 precision on a single NVIDIA H200 GPU with greedy decoding (no sampling, at most eight new tokens) and seed 20260901.

Every behavioral and mechanistic analysis used the same task prompt, which names both entities, defines the hazard as a spatial relation and restricts the answer to two words:

> "Inspect this construction scene for excavator struck-by risk. Is the visible worker inside or close enough to the excavator's operating, swing, travel, boom, or bucket zone that a struck-by hazard is present? Use only visible evidence. Answer with exactly one word: HAZARD or SAFE."

For each input, three outputs were recorded: the free-generated text, the 20 most probable tokens at the first decoding step with their log-probabilities, and a score for each candidate answer. Because the Qwen tokenizer encodes HAZARD as three tokens (HA, Z, ARD) and SAFE as one, candidate scores were computed over complete token sequences. For a candidate $c$ with tokens ($t_1$, …, $t_K$), the sequence log-probability is

$$\ell(c \mid x) = \sum_{j=1}^{K_c} \log p_\theta\left(t_j \mid x, t_{<j}\right) \tag{4}$$

and the HAZARD probability and decision score are

$$P_H(x) = \frac{\exp \ell(H|x)}{\exp \ell(H|x) + \exp \ell(S|x)}, \qquad S(x) = \log \frac{P_H(x)}{1 - P_H(x)} = \ell(H \mid x) - \ell(S \mid x) \tag{5}$$

where H and S denote HAZARD and SAFE and $p_\theta$ is the model's next-token distribution. Interventions on internal states require a score evaluated at a single position, so the causal and correction analyses used the first-step logit margin

$$m(x) = z_{\mathrm{HA}}(x) - z_{\mathrm{SAFE}}(x) \tag{6}$$

where $z$ denotes the pre-softmax logit of the first HAZARD subtoken or of SAFE at the first decoding step. The two scores were nearly identical for Qwen because the continuation tokens Z and ARD received a mean log-probability of −0.0005, and the free-generated label agreed with the label implied by Eq. (5) for 702 of the 710 controlled images (98.9%). For InternVL3.5, which generated HAZARD for every controlled image, labels were taken from its chat generation (448-pixel tiles, at most six tiles) and $m(x)$ served as the continuous score. The Qwen baseline and the co-occurrence controls used the processor's default image resolution, whereas activation extraction and interventions capped Qwen inputs at 256–512 × $28^2$ pixels to bound memory.

## 3.3 Dataset construction

### 3.3.1 Source data and scene selection

ConstructionSite-10k (Chen and Zou, 2026) was used as the primary source for struck-by hazards. It contains 10,013 construction-site photographs (7,009 training and 3,004 test) annotated for four safety rules. Its Rule 4, a worker inside the operating radius of an excavator, corresponds exactly to the struck-by relation studied here. Among its 10,013 photographs, 723 contained annotated instances of both a worker and an excavator and were retained as source scenes for counterfactual pair generation.

Two other datasets supplied object-level images: a construction-safety dataset with worker and protective-equipment annotations (LibreYOLO/construction-safety-gsnvb; 1,206 images) and an excavator dataset from the Roboflow-100 collection (2,655 images containing 1,530 excavator, 1,274 dump-truck and 1,080 wheel-loader instances). In addition, COCO 2017 (Lin et al., 2014; 37,003 training images) was used to complement the corpus for training sparse autoencoders.

To convert each candidate scene into a pair of images that differ only in the position of one worker, the six-step pipeline in Fig. 3 was implemented; Fig. 4 illustrates it for one scene. Source selection is described above and screening in Section 3.3.2; the intermediate steps were as follows.

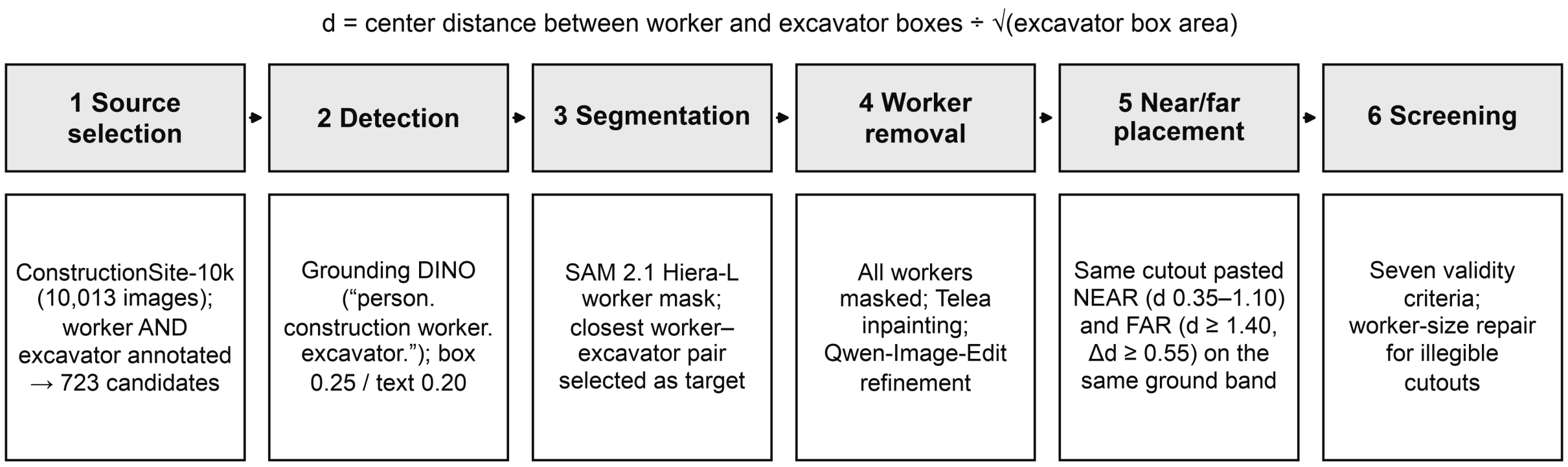

**Fig. 3.** Six-step pipeline for constructing the controlled counterfactual pairs, from source selection to screening; $d$ is the worker–excavator center distance divided by the square root of the excavator box area.

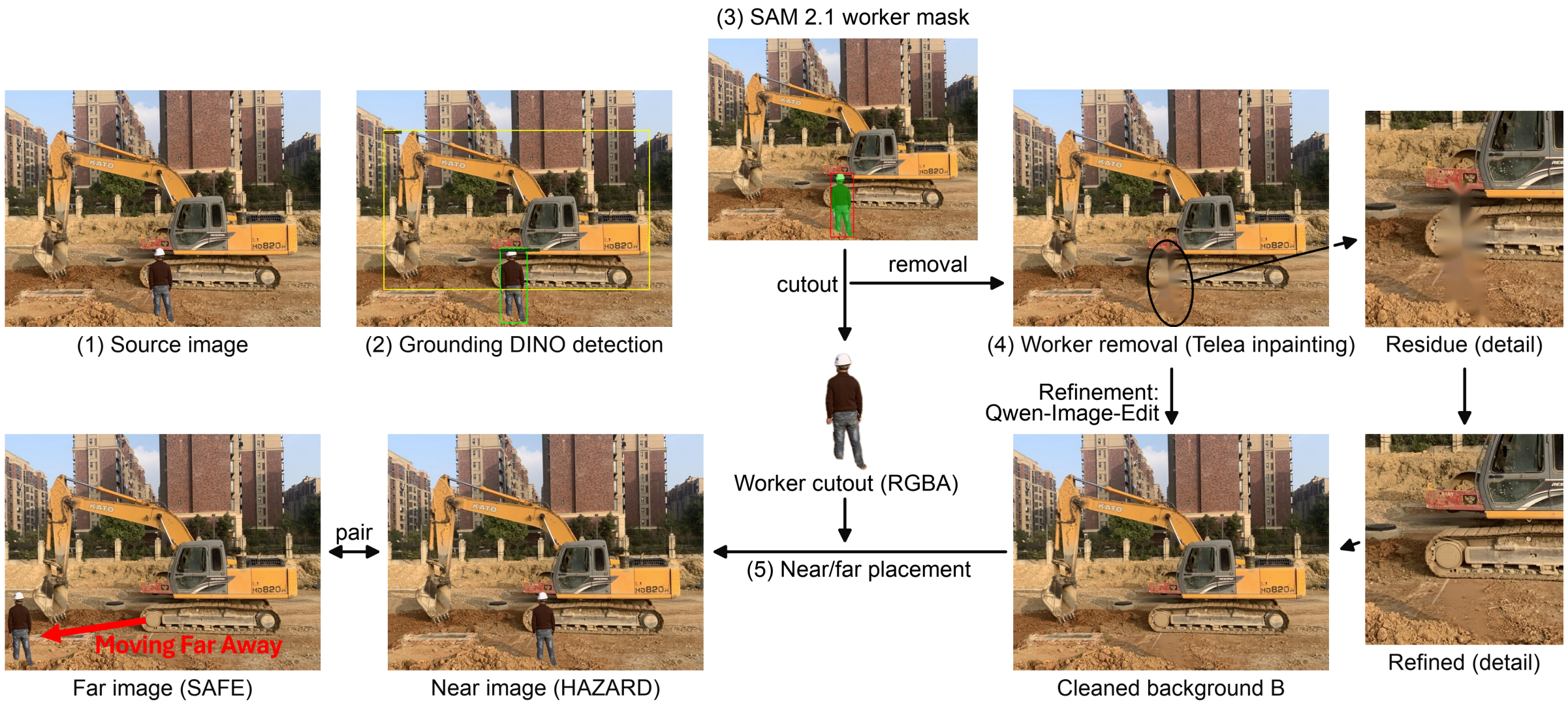

**Fig. 4.** Pair construction for one scene: (1) source image; (2) Grounding DINO detection of the worker and the excavator; (3) SAM 2.1 mask of the target worker, from which an RGBA cutout is taken; (4) removal of all workers by Telea inpainting, with the remaining residue refined by Qwen-Image-Edit to give the cleaned background $B$; (5) placement of the same cutout near the excavator (HAZARD) and far from it (SAFE). Details show the removal site before and after refinement.

*(1) Detection and target selection.* Workers and excavators were first localized with Grounding DINO (Liu et al., 2024) using the text query "person. construction worker. excavator." with box and text thresholds of 0.25 and 0.20, and duplicate boxes with an intersection-over-union above 0.55 were merged. Each worker box was converted into a pixel mask with SAM 2.1 (Hiera-Large; Ravi et al., 2025). A worker was eligible as the manipulation target only if its mask covered between 0.03% and 12% of the image, its bounding box measured at least 10 × 24 pixels with a height-to-width ratio of at least 1.15, and its padded crop did not touch the image border. The scale-free worker–excavator distance was defined as

$$d(w,e) = \frac{\|c_w - c_e\|_2}{\sqrt{A_e}} \tag{7}$$

where $c_w$ and $c_e$ are the centers of the worker and excavator bounding boxes and $A_e$ is the area of the excavator box. Normalizing by the square root of the excavator area expresses $d$ in excavator-size units, which makes it comparable across images of different resolution and camera distance. Among eligible workers, the target was the worker–excavator pair with the smallest $d$, with detection confidence breaking ties. The target was stored as an RGBA cutout whose alpha channel was smoothed with a Gaussian kernel ($\sigma = 1.2$ pixels).

*(2) Worker removal.* To ensure that each edited scene contained only one worker, all detected workers were first removed. Their masks were dilated with a 19-pixel elliptical kernel and filled by fast-marching inpainting (Telea, 2004). Because classical inpainting leaves smears on textured soil and machinery, the filled region was then refined with Qwen-Image-Edit (Wu et al., 2025; 50 denoising steps; true classifier-free guidance scale 4.0) applied to an expanded crop, using a prompt that requested the removal of human-shaped residue and the continuation of ground and machinery. The refined crop was feather-blended into the inpainted image, and every pixel outside the permitted write mask was copied unchanged. The result is referred to as the cleaned background, $B$.

*(3) Near and far placement.* The same cutout was composited onto $B$ at a near position $p_n$ and a far position $p_f$ chosen from a grid of candidate positions on rows aligned with the excavator's ground line. Let $d(p)$ be the distance of Eq. (7) for a worker placed at $p$, $o(p)$ the fraction of the worker box that overlaps the excavator box, $y(p)$ the vertical image coordinate and $H$ the image height. The admissible near and far sets were

$$\mathcal{P}_n = \{p\colon\ 0.35 \le d(p) \le 1.10,\ o(p) \le 0.12\} \tag{8}$$

$$\mathcal{P}_f = \{p\colon\ d(p) \ge 1.40,\ d(p) - d(p_n) \ge 0.55,\ |y(p) - y(p_n)| \le 0.12H\} \tag{9}$$

and the selected positions were

$$p_n = \underset{p\in\mathcal{P}_n}{\operatorname{argmin}} |d(p) - 0.72|, \qquad p_f = \underset{p\in\mathcal{P}_f}{\operatorname{argmax}} d(p) \tag{10}$$

The vertical constraint in Eq. (9) keeps the far worker on approximately the same ground band, and hence at a similar apparent depth, as the near worker. If either set was empty, a relaxed tier was used (near 0.25 $\leq d \leq 1.15$; far $d \geq 1.15$ with a separation of at least 0.40 and a vertical tolerance of $0.16H$). Near images were labeled HAZARD and far images SAFE.

*(4) Placement repair.* The first visual screening (Section 3.3.2) showed that many pasted workers were too small to identify, and the affected scenes were regenerated by re-pasting the original cutout at a legible scale. The cutout height was set to 9.5–14% of the image height (median 11%), the near position targeted $d \approx 0.78$, and the far position was required to satisfy $d(p_f) \geq d(p_n) + 0.40$, to lie at least 16% of the image diagonal from the near position, to remain within $0.10H$ of its ground band and to overlap the excavator by no more than 8%. A soft elliptical shadow was added beneath the feet.

### 3.3.2 Screening and final paired set

Successful generation does not guarantee that a pair isolates the intended relation. Every generated pair was subsequently screened manually against seven criteria: exactly one visible worker in each image; the worker fully visible and unclipped; identical worker appearance and scale in both images; clean removal at the original location; a near placement plausibly inside the operating, swing, travel, boom or bucket zone; a far placement plausibly outside it; and no alteration of machinery or scene geometry.

Initially, the pipeline was run until 500 pairs had been generated, which required processing 652 of the 723 candidate scenes. The 152 automatic failures arose because no sufficiently far position existed (115), no fully visible target worker was found (23), no near position existed (9) or no worker was detected (5). Of the 500 generated pairs, 80 passed the first screening. The 420 rejections were due to a worker too small to identify (315), residual workers (56), a clipped worker (36), visible editing artifacts (6), inconsistent worker appearance or scale (4) and implausible placement (3). The 315 pairs rejected for worker size entered placement repair, which produced 303 pairs (12 scenes had no admissible far position), of which 271 passed screening. This resulted in 355 pairs (710 images) in the final controlled set, with 80 first-pass and 271 repaired pairs. Near distances were tightly controlled (median $d = 0.79$), while far distances varied widely, ranging from workers just outside the zone to workers many machine lengths away (median 2.23; interquartile range 1.59–3.52; maximum 20.6; Fig. 5). Fig. 6 shows examples of accepted pairs, and Fig. 7 shows typical reasons for rejection at the first screening.

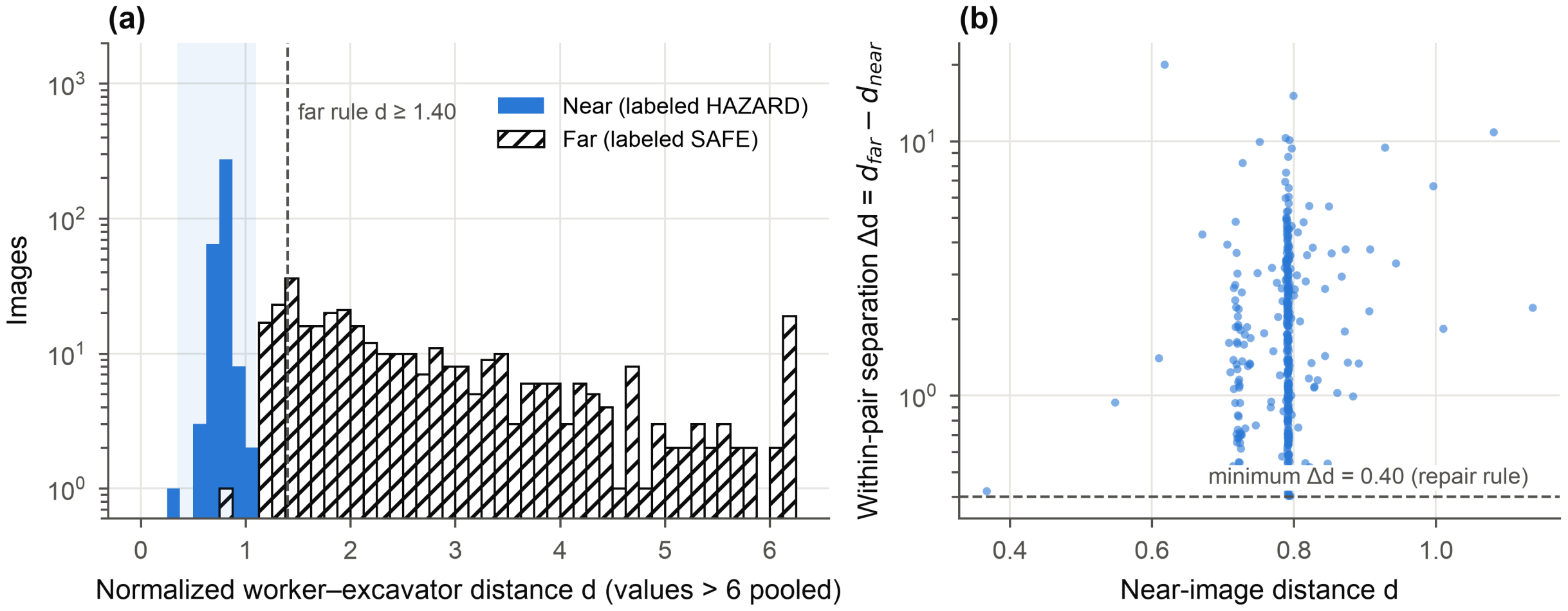

**Fig. 5.** Dataset characteristics. (a) Distribution of normalized worker–excavator distance for near and far images (log scale; values above 6 pooled). The shaded band marks the near placement rule and the dashed line the far rule. (b) Within-pair distance separation against near-image distance.

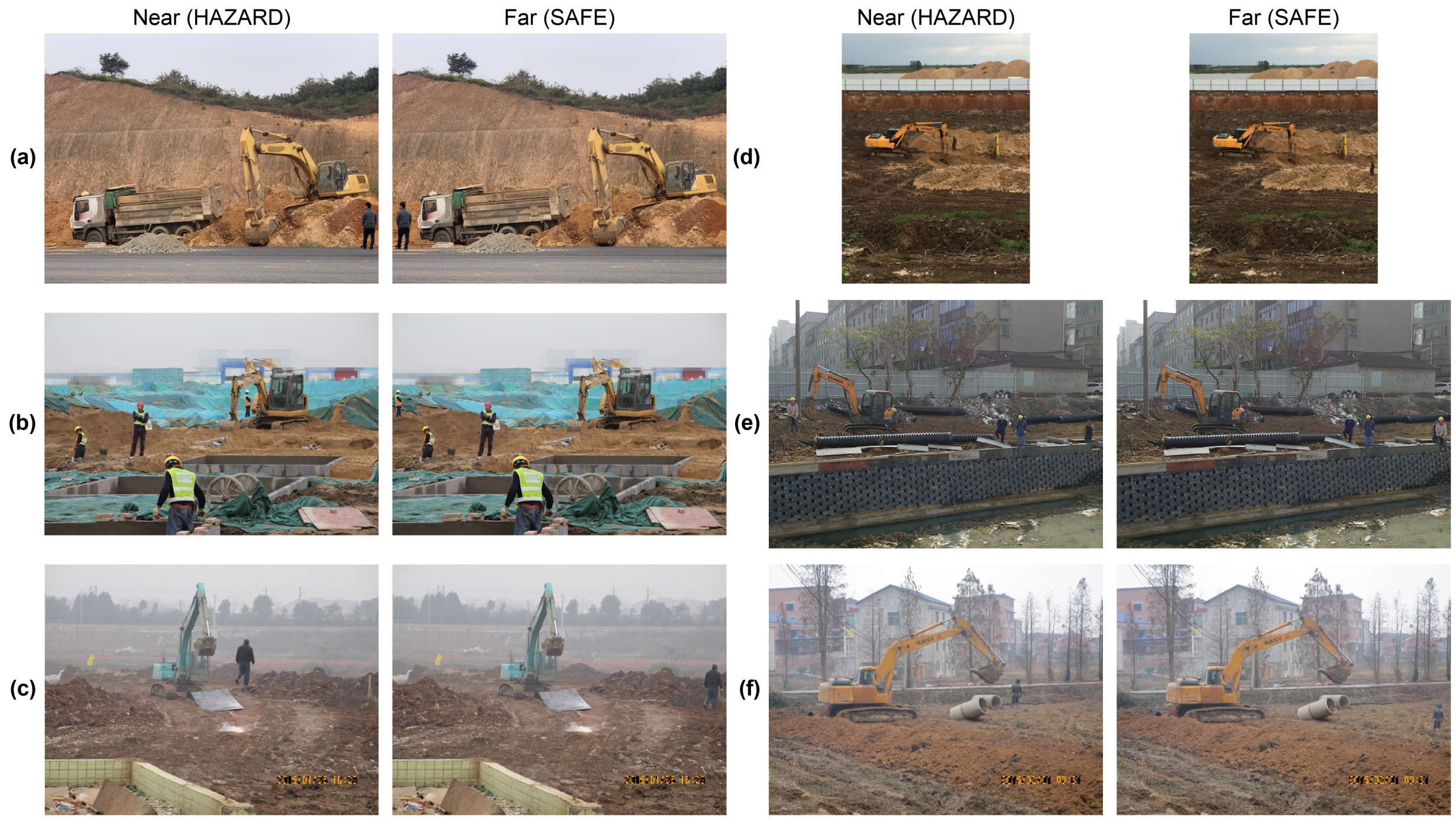

**Fig. 6.** Examples of accepted counterfactual pairs (a–f). In each pair the same worker cutout is placed near the excavator (labeled HAZARD) and far from it on approximately the same ground band (labeled SAFE); the rest of the image is the same cleaned background.

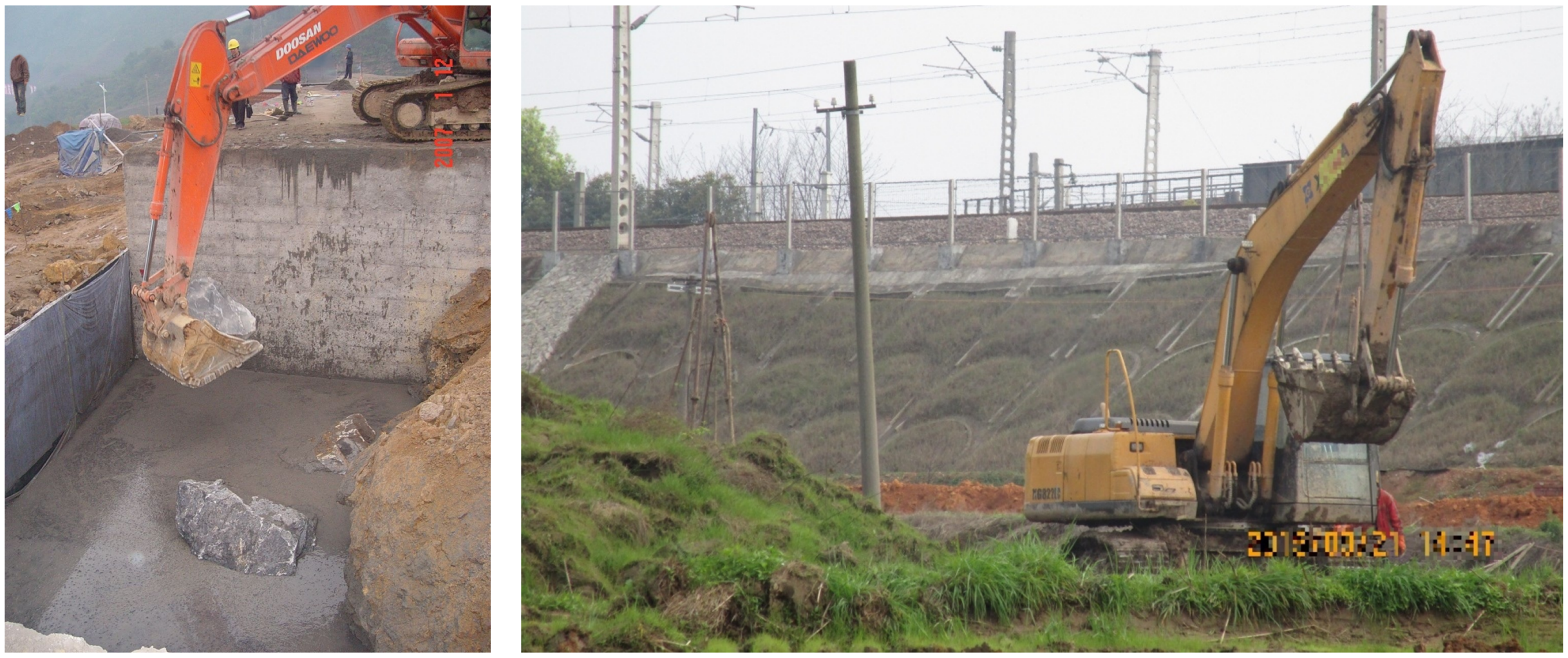

(a) Implausible placement (b) Clipped worker

**Fig. 7.** Examples of pairs rejected at the first visual screening: (a) implausible placement, in which the pasted worker does not stand on a plausible ground position; (b) clipped worker, in which the worker is not fully visible.

### 3.3.3 Control and transfer sets

The controlled pairs show whether the models encode and use the near/far relation, but three further questions cannot be answered from them alone: whether the hazard offset requires a worker to be present at all, whether the behavior observed on edited images also occurs in unedited photographs, and whether the internal representations can be analyzed with methods that need far more images than the 710 controlled images. Four additional image sets were constructed to answer these questions. The first, the set of worker-removed backgrounds, tests what produces the offset. If the offset arose from the scene and the prompt alone, a scene without any worker should score about as high as the far image; if it arises from the co-occurrence of the worker and the excavator, adding a worker anywhere should raise the score. The cleaned background *B* of each accepted pair provides this comparison, because it depicts the same scene, excavator and editing history as the pair but contains no worker. As these images were produced within the same pipeline as the counterfactuals, a difference between the score of *B* and that of the far image cannot be attributed to the editing procedure; it isolates the contribution of adding a worker (Section 3.4.3). The correct answer for every background is SAFE, because the prompt asks about “the visible worker”.

The second set, a four-condition natural set, serves two purposes. Probes for worker and excavator presence (Section 3.5.1) require labeled images in which each entity is present or absent, which the controlled pairs cannot supply because every pair contains both entities; and single-entity photographs show whether the co-occurrence effect observed on edited backgrounds also appears in images that were never edited. The set contains 200 photographs, 50 in each of four conditions: worker only (from the construction-safety dataset), excavator only (excavator annotations ranked first), worker and excavator far (Rule-4-negative ConstructionSite scenes ranked by distance; $d$ = 3.6–10.8, median 6.2) and worker and excavator near (Rule-4-positive scenes; $d$ = 0.07–0.99, median 0.53). It shares no source scenes with the edited pairs.

The third set, the Rule-4 natural transfer set, tests whether the findings obtained on edited pairs hold in unedited photographs whose labels reflect human judgment of the operating radius, which is the condition under which a deployed model would operate. This test is needed because edited images may carry compositing cues and because their near and far labels are defined by image-plane distance rather than by the operating radius judged on site. All 70 Rule-4-positive ConstructionSite images were combined with 70 negatives drawn from the 654 images that contain both an annotated worker and an annotated excavator without a Rule-4 violation. Because every negative contains both entities, co-occurrence is held constant across the two classes, so any discrimination between violations and compliant scenes must come from the spatial relation.

The fourth set supports the representation analysis. Sparse autoencoders learn a dictionary with several times more latents than activation dimensions, so they require far more training activations than the labeled sets provide, and they need varied images so that the learned features are not specific to the edited scenes. A corpus of 50,000 unlabelled images (all 10,013 ConstructionSite-10k images, 2,984 excavator-dataset images and 37,003 COCO 2017 training images; 44,907 for training and 5,093 for validation) was therefore assembled to train the sparse autoencoders (Section 3.5.3).

## 3.4 Behavioral evaluation

### 3.4.1 Classification and paired metrics

Let $\mathcal{N}$ and $\mathcal{F}$ denote the sets of near and far images, and let $N = 355$ be the number of pairs. Sensitivity, specificity and balanced accuracy are

$$\mathrm{Se} = \frac{1}{|\mathcal{N}|}\sum_{x\in\mathcal{N}} \hat{y}(x), \quad \mathrm{Sp} = \frac{1}{|\mathcal{F}|}\sum_{x\in\mathcal{F}}[1-\hat{y}(x)], \quad \mathrm{BA} = \frac{\mathrm{Se}+\mathrm{Sp}}{2} \tag{11}$$

Discrimination by the continuous score was summarized by the area under the receiver operating characteristic curve (AUROC), and binomial proportions are reported with Wilson 95% intervals. Because each scene contributes one near image $x_i^n$ and one far image $x_i^f$, the primary behavioral evidence is paired. The pairwise ordering rate and the mean paired shift are

$$\pi = \frac{1}{N}\sum_{i=1}^{N} \mathbb{1}\left[S(x_i^n) > S(x_i^f)\right], \quad \hat{\Delta} = \frac{1}{N}\sum_{i=1}^{N}\left[S(x_i^n) - S(x_i^f)\right] \tag{12}$$

A model that ignores the relation has $\pi = 0.5$ and $\hat{\Delta} = 0$. The paired shift was tested with a one-sided Wilcoxon signed-rank test, and each pair was classified as both correct, both HAZARD, both SAFE or reversed.

### 3.4.2 Distance response of false alarms

If false alarms on far images reflected genuine uncertainty about whether a nearby worker is exposed, their frequency should decline as the far worker moves away from the machine. Far images were therefore divided into quartiles of $d$, the HAZARD rate was estimated within each quartile, and the monotonic association between the far-image score and $d$ was measured with Spearman's rank correlation $\rho$.

### 3.4.3 Offset decomposition and co-occurrence increments

The terms of Eq. (2) were estimated by the mean far-image score, $\hat{b} = |\mathscr{F}|^{-1}$ Σ_{x∈$\mathscr{F}$} $S(x)$, and by $\hat{\Delta}$ from Eq. (12), which yields the PDI of Eq. (3). The worker-removed backgrounds provide a third observation of each scene under identical prompt, resolution and scoring, which allows $b$ to be partitioned. Using the first-step margin of Eq. (6), the co-occurrence increment and the relation increment of pair $i$ are

$$C_i = m(x_i^f) - m(B_i), \quad R_i = m(x_i^n) - m(x_i^f) \tag{13}$$

and their ratio is

$$\kappa = \frac{\overline{C}}{\overline{R}} = \frac{\sum_{i=1}^{N} C_i}{\sum_{i=1}^{N} R_i} \tag{14}$$

$C_i$ measures the hazard evidence added when a worker appears anywhere in the excavator's scene, and $R_i$ the evidence added when the same worker moves into the operating zone. A decision driven by the spatial relation implies $\kappa < 1$, whereas a co-occurrence override implies $\kappa > 1$. For Qwen, the background margins came from the control run and the near and far margins from the first-step log-probabilities of the baseline run, both at default resolution; for InternVL, all three margins came from the same runtime.

## 3.5 Representation analysis

### 3.5.1 Linear probes

Linear probes were used to test whether the entities and their relation are linearly recoverable from internal activations (Alain and Bengio, 2017). For Qwen2.5-VL, forward hooks recorded the activations of every vision-encoder block (32), the multimodal projector and every language-model layer (28). The activation of module $\ell$ for input $x$ was mean-pooled over its $T$ token positions,

$$\overline{h}_\ell(x) = \frac{1}{T}\sum_{t=1}^{T} h_{\ell,t}(x) \tag{15}$$

standardized with training-fold statistics, and passed to a logistic-regression probe

$$\hat{p}_\ell(x) = \sigma\left(w_\ell^\top \tilde{h}_\ell(x) + c_\ell\right) \tag{16}$$

where σ is the logistic function, $\tilde{h}$ is the standardized activation, and the weights w and bias $c$ were fitted by class-balanced, L2-regularized maximum likelihood. Worker-presence and excavator-presence probes were trained on 194 images of the four-condition set (six images flagged as rejection candidates were removed) with five stratified folds. The relation probe was trained on the 710 controlled images with five pair-grouped folds. If $\{\mathcal{G}_1, \ldots, \mathcal{G}_5\}$ partitions the 355 pairs and $\mathcal{J}_k$ contains both images of every pair in $\mathcal{G}_k$, the probe evaluated on $\mathcal{J}_k$ is fitted on the remaining images only, so that

$$\mathcal{J}_k \cap \mathcal{J}_{k'} = \emptyset \;\; (k \neq k'), \quad \left\{x_i^n, x_i^f\right\} \subseteq \mathcal{J}_k \;\; \forall i \in \mathcal{G}_k \tag{17}$$

Under this constraint, a probe cannot succeed by recognizing a scene it saw during training; it must generalize the relation to unseen scenes. Probe accuracy establishes that information is linearly available, not that the model uses it (Hewitt and Liang, 2019; Belinkov, 2022). Use was tested separately with interventions (Section 3.6).

### 3.5.2 Out-of-fold relation readout and error diagnosis taxonomy

For each model, the relation probe at the layer of peak decodability (language layer 21 for Qwen) was refitted with fixed pair-grouped folds, so that every image received an out-of-fold (OOF) prediction $q(x)$ from a probe that had never observed its pair; zero pair leakage was verified. For Qwen, OOF probes for worker presence (projector), excavator presence (vision layer 27) and a broader hazard proxy (language layer 0) were fitted with the same folds. Each behavioral error was then assigned to the earliest stage at which the OOF evidence failed. Writing $s_w$, $s_e$ and $s_r$ for the thresholded worker, excavator and relation readouts and $y$ for the true label, an error was labeled a perception error if $s_w = 0$ or $s_e = 0$; a relation error if both object signals were present but $s_r \neq y$; a reasoning error if all three signals were correct but the answer was wrong; and a decision error in the residual cases, in which the hazard-proxy readout and the answer disagreed. This taxonomy diagnoses where the available evidence diverges from the answer.

### 3.5.3 Sparse autoencoders

To test whether the entities and the relation are carried by individual interpretable features, top-$k$ sparse autoencoders (SAEs; Gao et al., 2025) were trained on activations from the projector, vision layer 27 and language layers 0, 15 and 21. For an activation $h \in \mathbb{R}^D$, the SAE computes

$$z = \mathrm{TopK}_k(W_{\mathrm{enc}}(h - b_{\mathrm{dec}}) + b_{\mathrm{enc}}), \quad \hat{h} = W_{\mathrm{dec}} z + b_{\mathrm{dec}} \tag{18}$$

where TopK retains the $k$ largest latent activations and sets the others to zero, and the parameters minimize the reconstruction error $\|h - \hat{h}\|^2$. The dictionary had $4D$ latents with $k = 32$ and was trained for 30 epochs (batch size 1,024, learning rate $3 \times 10^{-4}$, seed 42). Reconstruction quality was summarized by the fraction of variance explained, and latents were ranked on discovery data by oriented AUROC for worker, excavator, relation and hazard-proxy labels and evaluated on held-out data.

## 3.6 Causal analysis

Interventions tested which inputs and internal states the decision depends on. All were evaluated on the 355 controlled pairs with the first-step margin of Eq. (6). For an intervention $I$, the effect on the relational contrast is the change in the mean near/far gap,

$$\Delta G_I = \frac{1}{N} \sum_{i=1}^{N} \{[m_I(x_i^n) - m_I(x_i^f)] - [m(x_i^n) - m(x_i^f)]\} \tag{19}$$

where $m_I$ is the margin under the intervention. The changes in mean near and far margins were reported separately, because an intervention can narrow the gap by lowering near scores, by raising far scores or by shifting both.

*(1) Prompt-token ablation.* To measure how the words of the prompt contribute to the decision, the input embeddings $e_t$ of a word group $\mathcal{T}_g$ were set to zero while the sequence length and all other tokens were preserved:

$$e_t \leftarrow 0 \quad \text{for } t \in \mathcal{T}_g \tag{20}$$

The word groups were worker words ("visible worker"), excavator words ("excavator"), spatial-relation words ("inside", "close enough"), motion-zone words ("operating", "swing", "travel", "boom", "bucket", "zone"), the unions of these groups, and all task words. To provide length-matched controls, a neutral sentence ("This image is supplied as visual context for the assessment and no further written scene details are supplied.") was appended to the prompt in this analysis, and the same number of tokens was ablated within it.

*(2) Attention-head knockout and mediation.* For each of the 784 language attention heads (28 layers × 28 heads), the attention mass from the final prompt position to the image tokens was compared between near and far images with a paired Wilcoxon test, and $p$-values were adjusted with the Benjamini–Hochberg procedure (Benjamini and Hochberg, 1995). The eight highest-ranked heads were then removed as a group by zeroing their outputs before the attention output projection,

$$o_\ell = W_O^{(\ell)}\left[a_{\ell,1}; \dots; a_{\ell,H}\right], \qquad a_{\ell,h} \leftarrow 0 \;\; \forall(\ell, h) \in \mathcal{K} \tag{21}$$

where $a_{\ell,h}$ is the output of head $h$ in layer $\ell$, $H$ is the number of heads per layer and $\mathcal{K}$ is the knocked-out set. Eight heads from the same layers with low association served as matched controls. To test mediation, the outputs of the candidate heads, and separately of the control heads, were exchanged between the near and far images of each pair.

*(3) Residual-stream patching.* To locate where decision-relevant information resides, the residual state at layer $\ell$ and position set $\mathcal{P}$ (all image-token positions, or the final prompt position) was copied from the far image into the forward pass on the near image, and vice versa,

$$h_{\ell,t}(x_i^n) \leftarrow h_{\ell,t}\left(x_i^f\right) \;\; \forall t \in \mathcal{P} \tag{22}$$

and the resulting $\Delta G$ was recorded for every layer (Meng et al., 2022).

*(4) Object-recognition checks.* Yes/no prompts asking whether a worker or an excavator was visible were posed for all 710 controlled images, so that interventions could be checked for side effects on object recognition.

## 3.7 Correction analysis

The correction analysis asked whether the false-alarm tendency is implemented by identifiable model components, and whether the encoded relation can be used to correct the decision. All procedures used nested pair-grouped cross-validation: directions, layers, heads and neurons were selected on four folds and evaluated on the held-out fold.

### 3.7.1 Decodability and geometry of the false-alarm state

If false alarms reflect a distinct internal state, far images answered HAZARD should be separable from far images answered SAFE. OOF logistic probes of the form of Eq. (16) were trained at every language layer on the final-token residual state to discriminate far false alarms ($\mathscr{F}_{\mathrm{FP}}$) from far true negatives ($\mathscr{F}_{\mathrm{TN}}$). Within each training fold, a candidate false-alarm direction and a relation direction were defined as differences of class means,

$$v_C = \overline{h}(\mathcal{F}_{\mathrm{FP}}) - \overline{h}(\mathcal{F}_{\mathrm{TN}}), \qquad v_R = \overline{h}(\mathcal{N}) - \overline{h}(\mathcal{F}) \tag{23}$$

and their overlap was quantified by the cosine similarity $\cos(v_C, v_R)$ and by the loss of decodability of each label after the other direction had been projected out. The component of $v_C$ orthogonal to the relation direction is

$$v_C^{\perp} = v_C - \frac{v_C^{\top} v_R}{\|v_R\|_2^2} v_R \tag{24}$$

### 3.7.2 Directional steering

Steering tested whether moving the final-token state against the false-alarm direction clears far images without suppressing near images (Li et al., 2023a; Zou et al., 2023). At layer $\ell$, the final-token state was shifted as

$$h_{\ell,T} \leftarrow h_{\ell,T} - \alpha s_u \hat{u}, \qquad \hat{u} = \frac{u}{\|u\|_2} \tag{25}$$

where u is $v_C$ (raw steering) or $v_C^{\perp}$ (relation-preserving steering), $s_u$ is the training-fold standard deviation of the projections $h^{\top}\hat{u}$, which expresses the dose $\alpha$ in units of natural variation, and $T$ is the final prompt position. The layer and dose were chosen on 25 development pairs among layers 19–26, and effects were evaluated on the remaining 330 pairs against reversed, random, relation and answer-direction controls of equal norm.

### 3.7.3 Component ablation

Component ablation searched for attention heads and neurons whose removal selectively reduces false alarms. Initially, we ranked heads by their association with far false alarms and removed them with Eq. (21), but the selected heads overlapped with the relation-associated heads and with the data used to select them. Building on that, a new search procedure was built as follows. Each head's training-fold activation was correlated with the residualized far-image margin, from which the relation readout, the object readouts, the answer direction and the near/far indicator had been regressed out, and heads were scored as

$$\psi_j = \max\left(0, \rho_j^{\mathrm{far}}\right) + 0.5\left|\rho_j^{\mathrm{int}}\right| - 0.35\left|\rho_j^{\mathrm{rel}}\right| - 0.35\left|\rho_j^{\mathrm{gen}}\right| - 0.25\left|\rho_j^{\mathrm{near}}\right| \tag{26}$$

where $\rho^{\mathrm{far}}$ is the association with the residualized far margin, $\rho^{\mathrm{int}}$ the association with the far-by-answer interaction, and $\rho^{\mathrm{rel}}$, $\rho^{\mathrm{gen}}$ and $\rho^{\mathrm{near}}$ the associations with the relation readout, the generic answer direction and the near-image margin. Heads among the eight most associated with the relation or with the generic answer were excluded, and the eight highest-scoring remaining heads formed the candidate group. Ablation was applied at doses $\lambda \in \{0.25, 0.5, 0.75, 1\}$ as

$$a_{\ell,h} \leftarrow (1-\lambda)a_{\ell,h} + \lambda a^*_{\ell,h} \tag{27}$$

with $a^*$ equal to zero or to the training-fold mean activation. The candidate group was compared with relation heads, generic-answer heads, activation-matched and rank-matched controls, and 200 random head groups matched by layer. Individual neurons of the multilayer perceptrons were screened with the same residualization using a linear attribution score (weight × activation); the 64 highest-scoring neurons were pre-selected, and the 8 most stable per fold were ablated with Eq. (27).

A component was accepted as a localized false-alarm mechanism only if it met all criteria: a far-specificity gain of at least 0.05; a near-sensitivity loss of at most 0.02; retention of at least 90% of the relational gap; an AUROC loss of the relation readout of at most 0.02; an effect beyond the 95th percentile of the random groups; replication in at least four of five folds; and a double dissociation from the relation heads.

### 3.7.4 Relation-gated subspace steering

Finally, we tested whether the relation evidence that the model already encodes can condition a correction. Within each training fold, the standardized final-token residual states X at a candidate layer were residualized against the relation readout, the answer direction and the object readouts, and a rank-$r$ basis $U_r \in \mathbb{R}^{r\times D}$ ($r \in \{1, 2, 4, 8\}$) was obtained by truncated singular value decomposition. The steering direction was the unit vector within this subspace that best predicts the residualized far-image margin $\tilde{m}$,

$$\beta^* = \underset{\beta}{\operatorname{argmin}} \left\| \tilde{m} - XU_r^{\top}\beta \right\|_2^2, \qquad u_{\text{sel}} = \frac{U_r^{\top}\beta^*}{\|U_r^{\top}\beta^*\|_2} \tag{28}$$

and the layer and rank were chosen on training data to maximize the difference between the association of the projection with far-image margins and its absolute association with near-image margins. Steering followed Eq. (25) with $u = u_{\text{sel}}$ and $\alpha \in \{0.25, 0.5, 1, 2\}$, applied either to every image (ungated) or only when the OOF relation readout indicated a far placement (gated):

$$g(x) = \mathbb{1}[q(x) < 0.25], \qquad h_{\ell,T} \leftarrow h_{\ell,T} - g(x)\alpha s_u u_{\text{sel}} \tag{29}$$

where $q(x)$ is the OOF probability of a near placement given by the relation probe. Because the gate is computed by an external probe applied to the model's own activations, this procedure tests whether the encoded relation can correct the decision.

### 3.7.5 Benchmark against threshold calibration

Any intervention that lowers hazard scores can raise specificity at the cost of sensitivity, and so can a simple shift of the decision threshold. Each intervention was therefore compared with the operating curve obtained by varying $\tau$ in Eq. (1) on the unmodified score,

$$\mathcal{C} = \{ (\mathrm{Sp}(\tau),\ \mathrm{Se}(\tau)) \colon \tau \in \mathbb{R} \} \tag{30}$$

and it was regarded as a genuine improvement only if its (Sp, Se) point lay above $\mathcal{C}$, that is, if it achieved a higher specificity than threshold calibration at the same sensitivity.

# 4. Results

The results follow the diagnostic order of Fig. 1. Section 4.1 establishes the behavioral signature, Section 4.2 tests whether the entities and their relation are represented (H1, H2), Sections 4.3 and 4.4 quantify the relation-independent offset and identify what produces it (H3), Section 4.5 examines transfer to natural images, Section 4.6 locates where language and late states drive the decision, and Section 4.7 reports whether the encoded relation can correct the decision.

## 4.1 Answers move with the relation but remain hazard-saturated

Both models answered HAZARD for most images in which the worker stood outside the operating zone (Fig. 8a). Qwen2.5-VL labeled 300 of 355 near images HAZARD (sensitivity 84.5%) but cleared only 92 of 355 far images as SAFE (specificity 25.9%), for an accuracy of 55.2% and a score AUROC of 0.681. Its free-generated answers gave the same picture (near 84.8% HAZARD, far 76.1% HAZARD). InternVL3.5 answered HAZARD for all 710 controlled images (sensitivity 100%, specificity 0%), in every case as free-generated text rather than as an artifact of scoring.

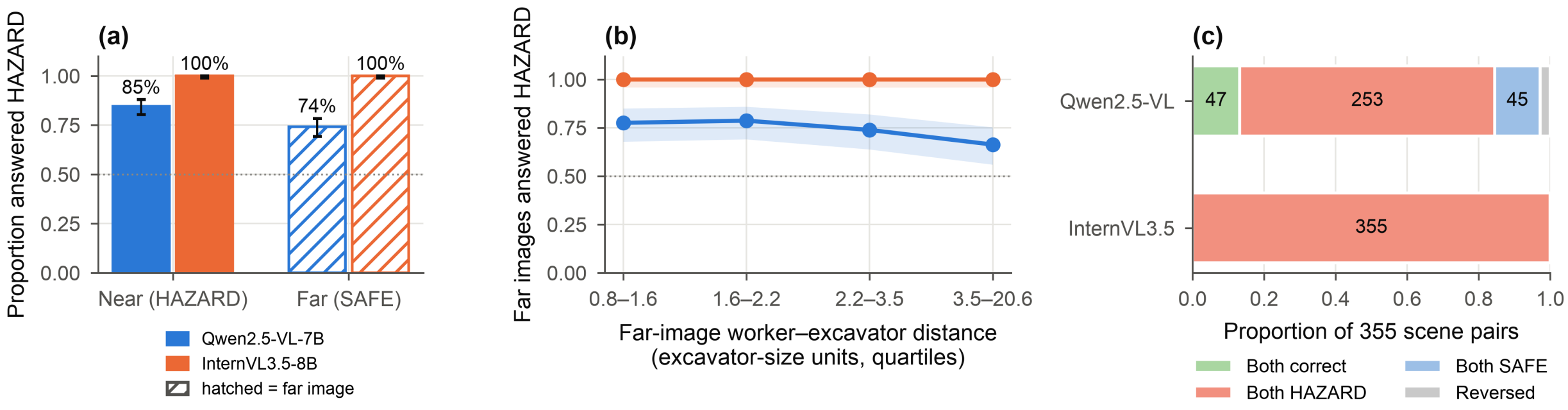

**Fig. 8.** Behavioral signature on the 355 controlled pairs. (a) Proportion of near and far images answered HAZARD with Wilson 95% intervals (Qwen2.5-VL: sequence-scored labels; InternVL3.5: free-generated

labels). (b) Far-image HAZARD rate across quartiles of worker–excavator distance; shading shows Wilson 95% intervals. (c) Pair outcome categories.

The answers nevertheless tracked the manipulated relation. For Qwen, the near image of a pair received the higher HAZARD probability in 78.6% of pairs, with a mean near-minus-far difference of 0.102 (95% CI 0.087–0.117; one-sided Wilcoxon $p = 8.4 \times 10^{-34}$). For InternVL, whose labels did not vary, the near image received the higher margin in 73.2% of pairs and the mean margin rose from 8.03 (far) to 9.06 (near). The pair outcomes show how the relational signal was lost at the decision (Fig. 8c): only 47 of 355 Qwen pairs were answered correctly on both images, 253 were answered HAZARD on both, 45 SAFE on both and 10 in reverse; all 355 InternVL pairs were answered HAZARD on both.

False alarms barely responded to how far the worker was from the machine (Fig. 8b). Across quartiles of far-image distance, Qwen's HAZARD rate was 77.5% ($d$ = 0.8–1.6), 78.7% (1.6–2.2), 73.9% (2.2–3.5) and 66.3% (3.5–20.6), and the far-image score was only weakly related to distance (Spearman $\rho = -0.10$, $p = 0.056$). InternVL answered HAZARD in every quartile; its margin declined slightly with distance ($\rho = -0.20$, $p = 1.2 \times 10^{-4}$) without approaching the threshold. A worker placed several machine lengths from the excavator was therefore still judged to be in danger in two of three Qwen cases and in every InternVL case.

## 4.2 The worker, the excavator and their relation are represented internally

Object identity was linearly available early and almost perfectly (Fig. 9a). In Qwen2.5-VL, worker presence was decodable with AUROC above 0.95 from about vision layer 8 onwards and reached 1.000 at the projector; excavator presence reached 0.999 at vision layer 27. When asked directly, Qwen confirmed an excavator in 98.2% of the 710 controlled images. It confirmed the worker less reliably (61.7%; 68.7% of near and 54.6% of far images), which reflects the smaller size of many pasted workers. A missed worker, however, removes the reason for a HAZARD answer, so it cannot explain the dominant false-alarm error.

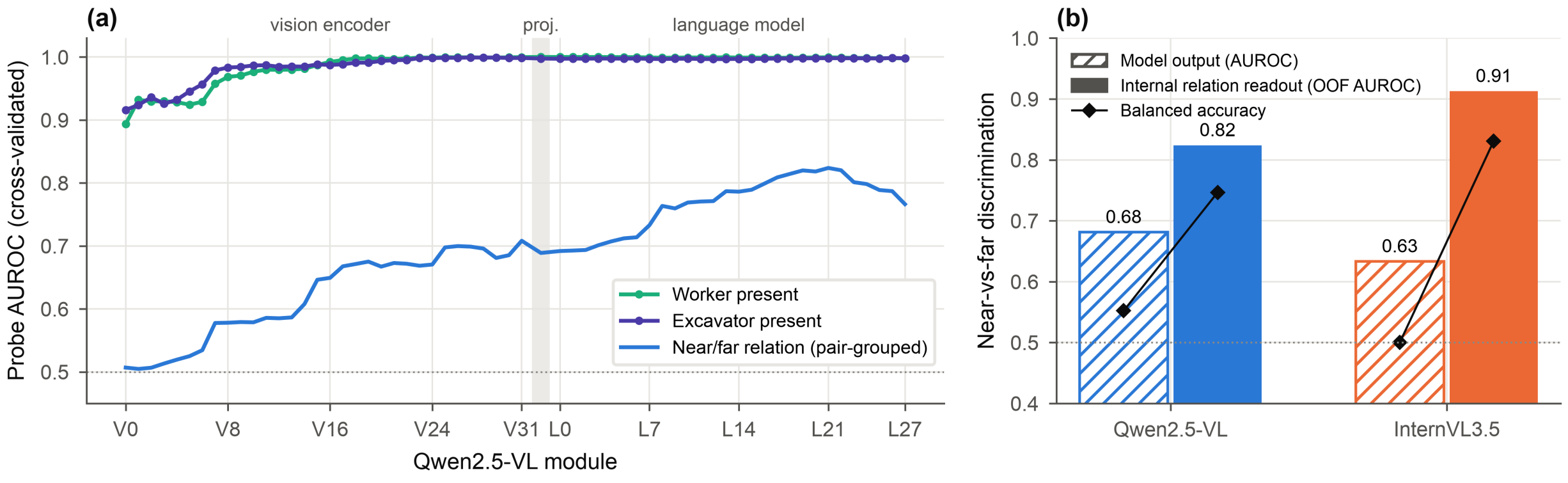

**Fig. 9.** The relation is available internally but under-used in the answer. (a) Cross-validated linear-probe AUROC across Qwen2.5-VL modules for worker presence, excavator presence and the controlled near/far relation (pair-grouped folds). (b) Near-versus-far discrimination by each model's answer score (hatched) and by an out-of-fold linear readout of its internal state (solid); diamonds give balanced accuracy.

The near/far relation, which is invisible to any single-object detector, was also decodable. Under pair-grouped validation, relation decodability rose slowly through the vision encoder (maximum AUROC ≈ 0.70) and continued to rise through the language model to a peak at layer 21 (AUROC 0.823), before declining in the last layers (Fig. 9a). The out-of-fold readout at that layer reached AUROC 0.824 (95% CI 0.799–0.846) and balanced accuracy 0.746 with zero pair leakage. In InternVL3.5 the corresponding readout reached AUROC 0.913 and balanced accuracy 0.831.

Fig. 9b contrasts these internal readouts with the models' own answers. Qwen's answer score separated near from far images with AUROC 0.681 and balanced accuracy 0.552, whereas a linear readout of its layer-21 state reached 0.824 and 0.746. For InternVL, the corresponding values were 0.633 and 0.500 for the answer against 0.913 and 0.831 for the readout. Both models therefore carried more information about the relation than their answers expressed. H1 and H2 cannot account for the errors: the entities are recognized, the relation is encoded and partially expressed in the score, and the shortfall arises after the relation is available.

Sparse autoencoders gave a consistent picture of object versus relation coding. Reconstruction explained 79.5–83.9% of activation variance across the five sites. Single latents tracked the objects (worker latent at the projector, held-out AUROC 0.857; excavator latent at vision layer 27, 0.826), but the best single latent for the controlled relation at language layer 21 reached only 0.548, indicating that the relation is carried by a distributed population code rather than by an identifiable feature.

## 4.3 A shared offset dominates the relational shift

Fig. 10 shows the score distributions of near and far images. In both models the far distribution lies mostly to the right of the decision threshold: 74.1% of Qwen far images and 100% of InternVL far images had a positive score. The relational shift moved near images further right but could not move far images across the threshold, because the far images started from a positive offset. For Qwen the offset was $b = 0.68$ log-odds and the relational shift $\Delta = 0.62$, giving a prior dominance index of 1.1 (95% CI 0.9–1.4). For InternVL, $b = 8.03$ and $\Delta = 1.03$ logit units, giving PDI = 7.8 (6.5–9.6). In Qwen the offset is comparable to the relational shift, so a third of the far images still fall on the SAFE side; in InternVL the offset is almost eight relational shifts, so no image does.

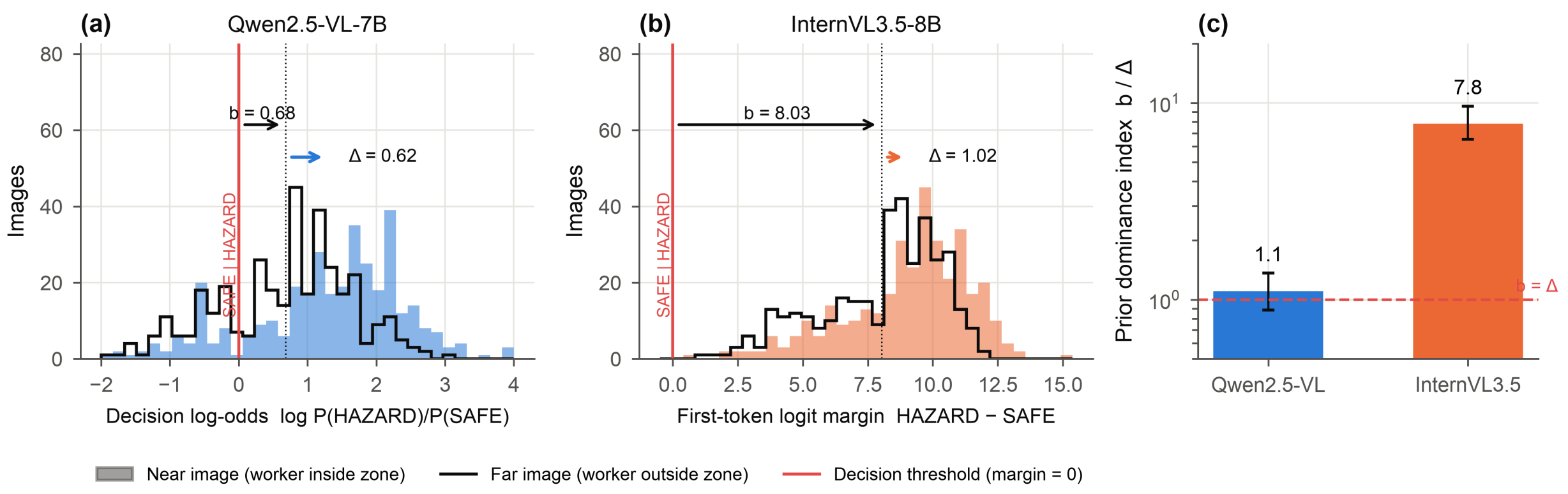

**Fig. 10.** Decision scores decompose into a large shared offset and a smaller relational shift. (a, b) Distributions of scores for near and far images; $b$ is the mean far-image score and $\Delta$ the mean within-pair near-minus-far shift. (c) Prior dominance index $b/\Delta$ with pair-bootstrap 95% intervals (log scale); values above 1 mean the offset exceeds the relational shift.

## 4.4 The offset is attached to the co-occurrence of worker and excavator

The worker-removed backgrounds and the four-condition natural set identify what produces the offset (Fig. 11). When the worker was removed from the same 355 scenes, Qwen answered HAZARD for only 11.0% of them, compared with 76.1% when the same worker was pasted far from the machine and 84.8% when it was pasted near. Qwen behaved the same way on natural images: 2% HAZARD for worker-only images, 34% for excavator-only images (23% after removing flagged images), 48% for natural far scenes and 98% for natural near scenes. Introducing a worker into an excavator scene thus changed Qwen's answer far more than relocating the worker relative to the operating zone.

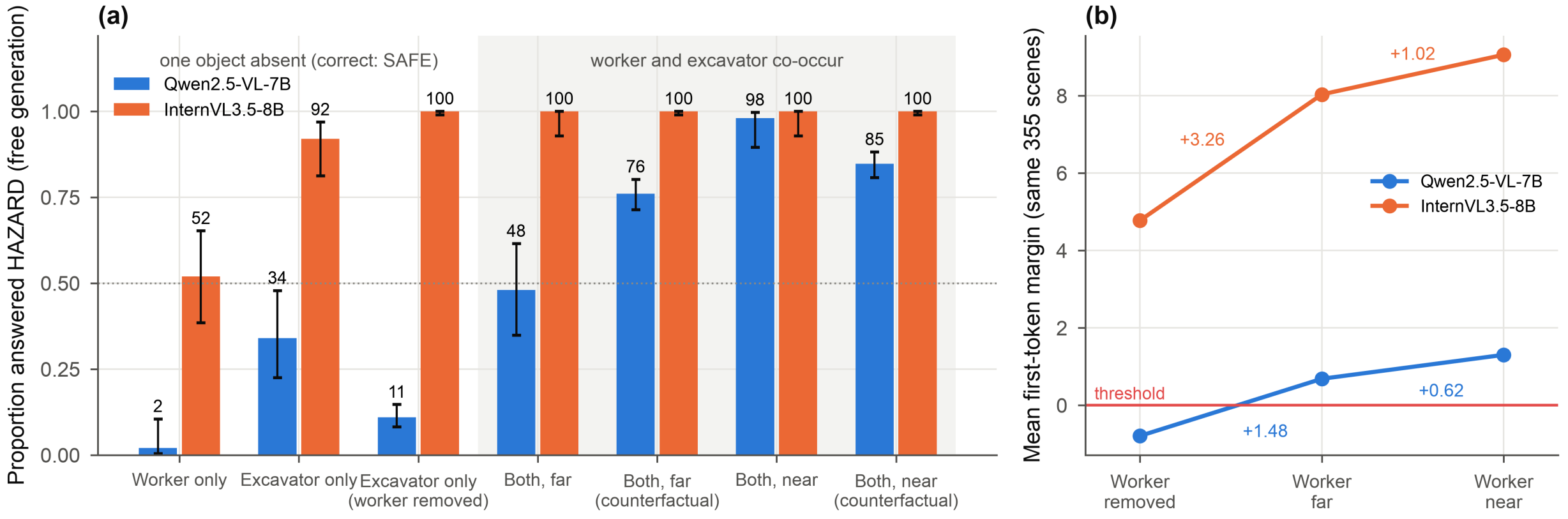

**Fig. 11.** The hazard offset is attached to worker–excavator co-occurrence. (a) Proportion of images answered HAZARD (free generation) when one entity is absent (natural worker-only and excavator-only images; the same 355 scenes with the worker removed) and when both co-occur (natural and counterfactual far and near images); Wilson 95% intervals. (b) Mean first-step margin for the same 355 scenes with the worker removed, placed far and placed near; labels give the co-occurrence increment (removed → far) and the relation increment (far → near).

The pair-level decomposition makes this precise (Fig. 11b). For Qwen, adding a worker far from the excavator raised the first-step margin by $C = 1.48$ (95% CI 1.38–1.57), whereas moving the same worker into the zone raised it by a further $R = 0.62$ (0.54–0.70). The co-occurrence increment was 2.4 times the relation increment (2.1–2.8), the far image scored above its own background in 94.4% of pairs, and $C$ exceeded $R$ in 76.3% of pairs. InternVL showed the same ordering at a larger scale: $C = 3.26$ (2.97–3.54), $R = 1.03$ (0.84–1.22), a ratio of 3.2 (2.6–4.0), with the far image above its background in 83.1% of pairs.

The two models differ in where the offset begins. In Qwen the scene and prompt alone produce a negative margin (−0.79; 11% HAZARD), and it is the arrival of a worker that pushes the score across the threshold. In InternVL the worker-removed scene already scores 4.77 and is answered HAZARD in every case; natural excavator-only images are answered HAZARD in 92% of cases and even worker-only images in 52%. For InternVL, then, an excavator in view combined with a prompt that names a worker and an excavator is sufficient for a HAZARD answer, and the worker's presence and position add to an offset that is already decisive. In both models the hierarchy of evidence is the same: the presence of the second entity contributes two to three times more to the hazard score than the spatial relation that defines the hazard.

### 4.5 Object and co-occurrence signals transfer to natural scenes; the controlled relation readout does not

On the balanced Rule-4 set, in which every safe image also contains a worker and an excavator, Qwen labeled 65 of 70 violations HAZARD (sensitivity 92.9%) but cleared only 8 of 70 compliant scenes (specificity 11.4%), and its score separated the two classes barely better than chance (AUROC 0.557; Fig. 12b). Because both classes contain the two entities, the co-occurrence offset of Eq. (2) is shared by violations and compliant scenes and cannot discriminate between them, so any discrimination must come from the relational term. The near-chance AUROC combined with the extreme imbalance between sensitivity and specificity is the pattern expected when a large, class-independent offset is added to a weak relational signal: almost every image crosses the threshold, and the remaining ranking carries little information about exposure. The co-occurrence-driven false alarm observed on edited pairs therefore reappeared in unedited photographs. This result also shows how the failure can be concealed: in benchmarks in which many safe images lack one of the entities, a model can score well by detecting co-occurrence alone, whereas here, with both entities present in every safe image, the same model performed near chance.

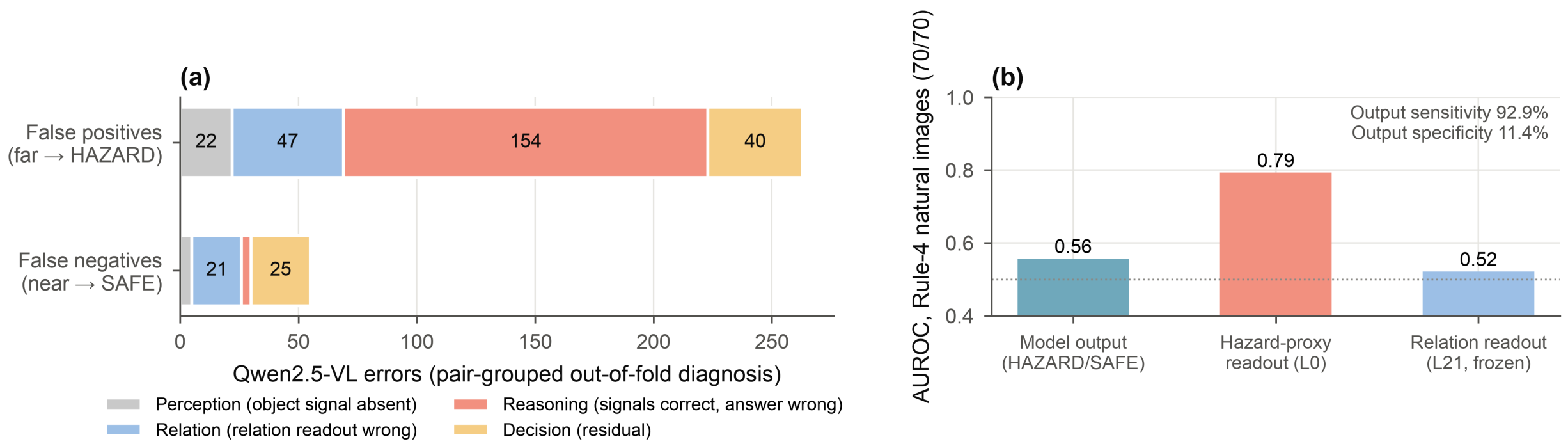

**Fig. 12.** Error diagnosis and natural transfer (Qwen2.5-VL). (a) Pair-grouped out-of-fold assignment of false negatives and false positives to the earliest failing stage. (b) AUROC on the balanced Rule-4 natural set for the model's answer score, a frozen hazard-proxy readout and the frozen controlled relation readout.

The frozen readouts indicate which kind of evidence generalizes. Object probes transferred without loss, detecting a worker in 96.4% and an excavator in 100% of the natural images, so the entities remained recognized outside the edited domain. A broader hazard-proxy readout from language layer 0 retained an AUROC of 0.793. Because this layer still largely reflects the visual input and the scene context passed through the projector, the readout captures whether an image looks like a hazardous construction scene, and such scene-level evidence transferred from the four-condition natural set to Rule-4 photographs. The controlled relation readout, in contrast, did not transfer (AUROC 0.521, balanced accuracy 0.514). It was fitted on edited pairs in which a single, upright, cleanly segmented worker moved along one ground band, and it did not survive the change to natural scenes with several workers, varied viewpoints, partial occlusion and an operating radius annotated by human judgment rather than by image-plane distance. The evidence that drives the offset is therefore robust and general, whereas the evidence that should correct it is fragile and specific to the domain in which it was learned.

## 4.6 Language and late decision states carry the offset

The out-of-fold error taxonomy located most Qwen false alarms after the relation stage (Fig. 12a). Of the 263 far-image false alarms, 154 (58.6%) occurred although the object signals were present and the relation readout correctly indicated a far placement (reasoning stage), 47 (17.9%) coincided with an incorrect relation readout, 40 (15.2%) were assigned to the decision stage and 22 (8.4%) to perception. The 55 missed hazards were distributed differently: 25 decision, 21 relation, 5 perception and 4 reasoning. False alarms were thus mostly cases in which the model had the evidence needed for a SAFE answer and did not use it.

Prompt-token ablation showed that the words naming the entities and the relation shape both the offset and the gap (Fig. 13; this run used the capped-resolution pipeline with an appended neutral sentence, under which the baseline far HAZARD rate was 88.2%). Zeroing any group of task words raised far-image margins and narrowed the near/far gap. Removing the worker words left the far HAZARD rate at 98.6% while shrinking the gap from 0.54 to 0.20; removing the worker and excavator words together drove all far images to HAZARD (gap 0.13); removing all task words raised the mean far margin from 1.03 to 4.15 and nearly eliminated the gap (0.07). Removing the excavator word alone raised both margins and slightly widened the gap (0.60). Length-matched ablations of the neutral sentence changed far margins by −0.29 to +0.31 and left the gap between 0.53 and 0.66. With the descriptive words of the prompt removed, the model fell back on a generic HAZARD answer that no longer depended on the image; the words that define the relation were needed to express the relational difference at all, but they did not remove the offset.

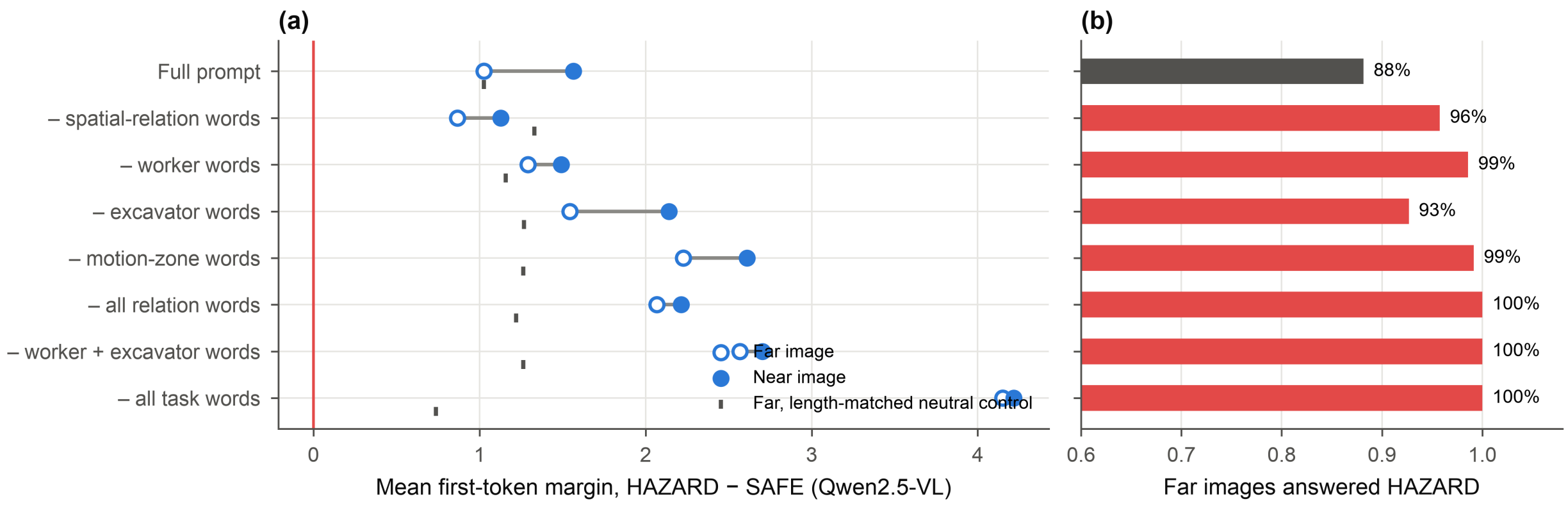

**Fig. 13.** Prompt-token ablation (Qwen2.5-VL, capped-resolution pipeline). (a) Mean first-step margin for near (filled) and far (open) images after zeroing the embeddings of each word group; ticks show far-image margins under length-matched ablation of a neutral sentence. (b) Far images answered HAZARD under each ablation.

Causal tests located the decision-relevant information in late language states instead of in a compact circuit (Fig. 14). Of the 784 attention heads, 419 differed in their attention to image tokens between near and far images after false-discovery-rate correction, with the strongest effects in layers 19–24. Knocking out the eight top-ranked heads reduced the near/far gap by 0.061 (95% CI 0.044–0.076; $p = 6.3 \times 10^{-12}$), about 11% of the baseline gap, mainly by lowering near-image margins (−0.055) while leaving far margins unchanged (+0.005); eight matched control heads changed the gap by +0.001. Cross-patching the candidate heads, however, was no more effective than patching the controls (gap change −0.024 versus −0.029), so the heads contribute to the relational shift without uniquely mediating it, and neither head group reduced the offset. Exchanging residual states between near and far images at the final prompt position reversed the gap at every late layer, with the largest effect at layer 26 (−1.15; 95% CI −1.30 to −1.00, about twice the baseline gap), showing that late final-token states carry the complete decision-relevant contrast.

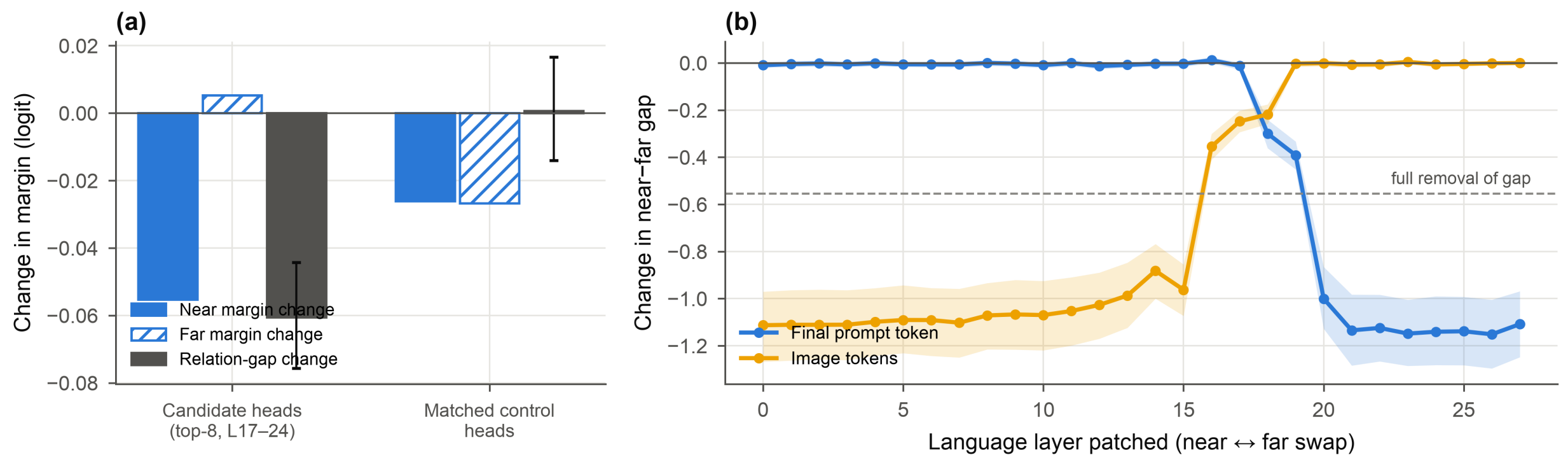

**Fig. 14.** Causal localization (Qwen2.5-VL). (a) Change in near margin, far margin and near/far gap after knocking out the top-eight relation-associated attention heads or eight matched control heads (95% interval for the candidate gap change). (b) Change in the near/far gap when residual states are exchanged between near and far images at each language layer, at the final prompt position or the image-token positions; shading shows pair-bootstrap 95% intervals and the dashed line the change that would exactly remove the baseline gap.

## 4.7 Model-internal components do not remove the offset; routing the encoded relation into the decision partly does

Table 1 and Fig. 15 summarize the correction analyses against plain threshold calibration. In Qwen, far false alarms were decodable from far true negatives out of fold, peaking at language layer 20 (AUROC 0.795, 95% CI 0.746–0.841). The false-alarm direction overlapped partly with the relation direction (mean cosine 0.22), and removing either direction reduced decoding of the other only slightly. Steering along the false-alarm direction, raw or orthogonalized to the relation direction, changed far specificity by +0.003 and near sensitivity by −0.006 or +0.006 on 330 held-out pairs. The initial head search raised far specificity by 0.149, but at a cost of 0.101 in near sensitivity, because the selected heads lowered near and far margins almost equally (−0.70 and −0.64); its sensitivity–specificity point lay on the threshold-calibration curve, and the heads overlapped with relation components. The revised, overlap-free head ablation met only three of eight criteria in Qwen (far specificity −0.056; random-group $p = 0.975$) and three of eight in InternVL (no change in either rate; relation gap retained 47%, although the effect exceeded random groups, $p = 0.005$). No single neuron passed the criteria in either model; the 101 neuron interventions tested in Qwen changed far specificity by −0.048 to +0.051 and near sensitivity by −0.025 to +0.031, clustering around the calibration curve.

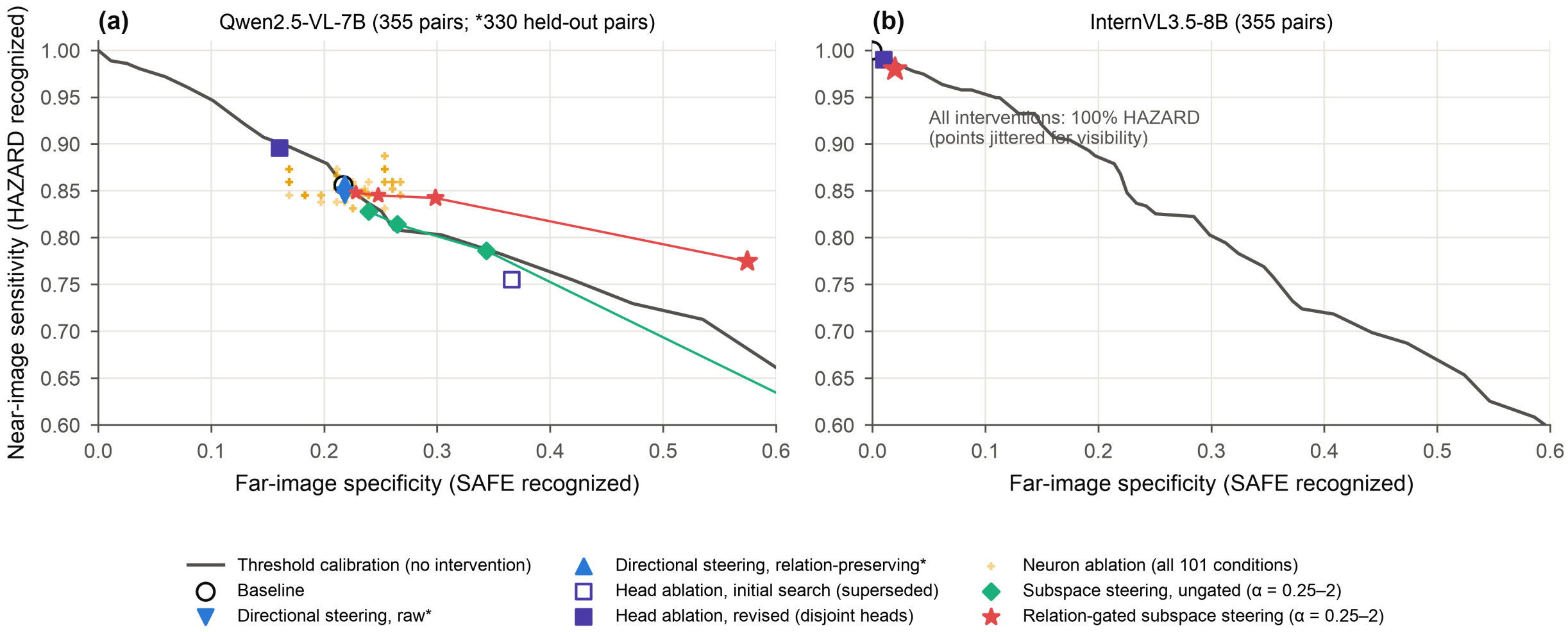

**Fig. 15.** Correction analyses in the sensitivity–specificity plane. The gray curve shows what threshold recalibration alone achieves on the unmodified score. Points show held-out results of each intervention; neuron ablation is shown for all 101 single-neuron conditions and subspace steering for the full dose series ($\alpha = 0.25$–$2$). Only the subspace intervention gated by the model's own relation readout lies above the calibration curve (Qwen2.5-VL); no InternVL3.5 intervention changed any answer (points jittered for visibility).

Only one manipulation moved the operating point above the calibration curve: steering along the selective subspace only when the model's own relation readout indicated a far placement (relation-gated subspace steering). In Qwen, the gated intervention raised far specificity from 0.217 to 0.299 at $\alpha = 1$ with a sensitivity loss of 0.014, and to 0.575 at $\alpha = 2$ with sensitivity 0.775. At the latter sensitivity, threshold calibration reaches a specificity of only 0.355, and the gated intervention's balanced accuracy (0.675) exceeded the best value attainable by any threshold (0.648). The same subspace applied without the gate followed the calibration curve. In InternVL, no tested dose changed any answer, because doses scaled to the activation spread could not overcome an offset of about eight logits. These results answer RQ4 in two parts. No internal component was found whose removal selectively cleared far images, so the offset is not localized in a dedicated circuit. The decision can, however, be corrected when the relation evidence the model already encodes is used to condition the intervention, and the size of the correction is limited by the size of the offset.

**Table 1.** Correction analyses (held-out evaluation; Qwen2.5-VL unless noted). Baseline in the capped-resolution intervention pipeline: near sensitivity 0.856, far specificity 0.217.

| **Analysis** | **Intervention** | **Δ far specificity** | **Δ near sensitivity** | **Above calibration curve?** | **Localization criteria met** |
|---|---|---|---|---|---|
| False-alarm decoding | Probe far false alarm vs true negative (L20) | AUROC 0.795 (decodable) | – | – | – |
| Direction geometry | Cosine of false-alarm and relation directions | 0.22 | – | – | partial overlap |
| Directional steering, raw (330 pairs) | Steer against $v_C$ | +0.003 | −0.006 | No | No |
| Directional steering, relation-preserving (330 pairs) | Steer against $v_{C\perp}$ | +0.003 | +0.006 | No | No |
| Head ablation, initial search (superseded) | False-alarm-associated heads | +0.149 | −0.101 | No (on curve) | No (overlap) |
| Head ablation, revised | Disjoint residualized heads | −0.056 | +0.039 | No | 3/8 (Qwen); 3/8 (InternVL) |
| Neuron ablation (101 conditions) | Single neurons | −0.048 to +0.051 | −0.025 to +0.031 | No | No |
| Subspace steering, ungated ($\alpha = 2$) | Selective subspace on all images | +0.484 | −0.282 | Marginal | – |
| Relation-gated subspace steering (α = 1) | Selective subspace when readout indicates far | +0.082 | −0.014 | Yes | external gate |
| Relation-gated subspace steering (α = 2) | Selective subspace when readout indicates far | +0.358 | −0.082 | Yes (+0.22 at equal sensitivity) | external gate |
| All subspace conditions (InternVL) | – | 0.000 | 0.000 | No | No |

# 5. Discussion

## 5.1 Why VLMs over-report struck-by hazards

The three explanations in Fig. 1 make different predictions, and the evidence favors the third. Perception failure (H1) predicts weak object representations and an excess of missed hazards. Instead, both objects were decodable almost perfectly, excavators were confirmed in 98% of images, and 263 of Qwen's 318 errors were false alarms. Relation failure (H2) predicts that the near/far relation cannot be read from the model and that answers ignore worker movement. Instead, the relation was decodable out of fold (AUROC 0.82 and 0.91), and the answer score ranked the near image above the far image in about three of four pairs. Co-occurrence override (H3) predicts that the relation is available but outweighed by an offset tied to the presence of both entities, and three independent observations confirm it. Far-image false alarms changed little across a 25-fold range of worker–excavator distance; removing the worker from the same scenes reduced Qwen's free-generated HAZARD rate from 76% to 11%; and, across the same 355 scenes, the hazard evidence added by a worker's presence exceeded the evidence added by the worker's position by factors of 2.4 (Qwen) and 3.2 (InternVL). The error taxonomy pointed the same way: most false alarms occurred when the model's own internal signals correctly indicated a distant worker.

The co-occurrence account extends findings on object hallucination and language priors (Rohrbach et al., 2018; Li et al., 2023b; Zhou et al., 2024) from object presence to a relational safety decision. In hallucination studies, co-occurrence statistics cause a model to report an absent object; here, both objects are present and the statistic attached to their joint presence determines the answer to a question about their arrangement. The account also qualifies the common view that VLMs fail spatial tasks because they do not represent spatial relations (Kamath et al., 2023; Yuksekgonul et al., 2023). In this task the relation was represented, partly expressed in the score and carried into late language layers, yet it was outweighed at the decision. Benchmark accuracy would have attributed both models' failures to poor spatial understanding; the internal evidence however shows a calibration problem layered on top of adequate spatial encoding.

These results are consistent with recent mechanistic studies of spatial reasoning in VLMs. Some of these studies attribute spatial failures to deficits in encoding, such as attention that fails to align with the true object locations (Chen et al., 2025) or vision-token embeddings whose large norms swamp positional information in the language model (Qi et al., 2026). Others suggest that spatial information can be decodable without being used: object positions become decodable before the relation decision emerges (Liu et al., 2026a), and encoded spatial directions influence answer logits only at mid-to-deep layers, with positive effects that can remain below the decision threshold (Wang and Xu, 2026). The present findings support the second account for struck-by hazards and identify what outweighs the encoded relation: the evidence attached to the co-occurrence of the worker and the machine. Probes of VLM embeddings have also shown that perceived distance is entangled with vertical image position (Min et al., 2026). Because the far worker in each pair was kept on approximately the same ground band as the near worker (Eq. (9)), the relation readouts reported here are unlikely to reflect vertical position alone.

In addition, the two VLM models illustrate two strengths of the same failure. In Qwen the scene and prompt alone produced a SAFE-leaning score, and the arrival of a worker, wherever placed, tipped the decision. The offset was comparable to the relational shift (PDI 1.1), so a minority of far images were still cleared. In InternVL the excavator and the prompt together were already sufficient for a HAZARD answer, even on worker-removed scenes, and the offset was nearly eight relational shifts, so no image was cleared. InternVL also encoded the relation more strongly than Qwen. Better relation encoding therefore does not by itself yield better decisions; the balance between offset and relational shift does.

The token-ablation results indicate how language contributes. The prompt names the worker, the excavator and the hazard; when the words that specify the relation or the entities were removed, far-image margins rose and the near/far gap shrank, and with all task words removed the model answered HAZARD for every image irrespective of content. The descriptive words of the prompt are thus what allows the relation to be expressed at all; without them the model defaults to a generic hazard answer. The hazard concept appears to be evoked by the construction context and the hazard framing, while the relational words provide a comparatively weak modulation. This is consistent with the observation that the hazard-proxy readout in early language layers transferred to natural images whereas the controlled relation readout did not.

## 5.2 Does the manipulation analysis resolve the problem?

The correction analyses answer this question in two parts. First, they did not find a localized internal mechanism whose removal clears far images. Directions that decode false alarms could be found (AUROC 0.80), but steering along them had negligible effect; a head group that appeared to reduce false alarms in an early analysis did so by lowering all hazard scores, which threshold recalibration achieves equally; and the revised, overlap-free head and neuron searches failed their prespecified criteria in both models. The offset thus behaves as a distributed property of how the model maps construction scenes and hazard prompts to answers, not as a removable module. A negative ablation result must nevertheless be interpreted with care, because visual information can reach the answer through more than one route and models can switch routes when one is ablated (Salazar et al., 2026); ablation reveals what a model can do without a component rather than what it normally does. The failure of every localization criterion is consistent with such redundancy and with a distributed offset rather than a dedicated circuit. This negative result is informative for practice: approaches that seek to patch a single component are unlikely to fix the problem.

Second, the analysis shows that the problem can be partly corrected when the decision is conditioned on the relation evidence the model already encodes. Steering applied only to images whose internal relation readout indicated a far placement moved Qwen's operating point above the threshold-calibration curve, raising far-image specificity from 0.36 to 0.58 at equal sensitivity (0.78) and lifting balanced accuracy beyond the best threshold. The same intervention without the gate did not beat calibration. The gain therefore comes from using the encoded relation, which is what the co-occurrence account predicts: the model has the evidence but does not weigh it. The correction remained partial. The gate relies on an external probe trained on the controlled pairs, the readout did not transfer to natural images, and in InternVL the tested doses were too small relative to the offset to change any answer. Simply reading the relation from the internal state, without steering, achieved balanced accuracies of 0.75 (Qwen) and 0.83 (InternVL), higher than any steered or recalibrated output, which marks how much of the problem lies between representation and answer.

In summary, although the manipulation analyses do not yet resolve the problem in a completely deployable form, they point out a valuable direction: the remedy must strengthen the influence of encoded relational evidence on the decision, or weaken the co-occurrence offset, rather than improve object perception or add spatial capacity that the models already have.

## 5.3 Implications for construction safety practice

Three implications follow for practitioners and developers who intend to use VLMs for hazard monitoring. First, high sensitivity should not be read as competence. Both models detected most genuine hazards, but they did so largely because they called almost every worker near plant hazardous; on natural Rule-4 images, Qwen's 93% sensitivity came with 11% specificity. In a monitoring system this would produce alerts for most routine work near machines, a recognized cause of alarm fatigue and of operators disabling warnings. Evaluations should therefore report specificity on images in which the hazard-relevant entities co-occur safely, not only accuracy on mixed sets in which many safe images lack one of the entities.

Second, evaluation datasets for relation-defined hazards should include matched counterfactuals. Natural datasets confound relation with scene, viewpoint and worker count, and a model can score well on them by recognizing hazardous-looking scenes. The near/far and worker-removed controls used here expose this behavior directly and can be built from existing annotated site images with open detection, segmentation and editing tools. The screening record shows, however, that most automatically generated pairs fail visual validity checks, so manual or verified screening is essential.

Third, prompt engineering and threshold tuning alone are unlikely to be sufficient. Thresholds can trade sensitivity for specificity along a fixed curve, but cannot improve the curve; the offset reflects learned associations rather than wording, as the token ablations showed. More promising strategies are those that make the decision depend on explicit relational evidence: hybrid pipelines in which a geometric module or a relation readout gates the VLM's answer, fine-tuning on balanced counterfactual pairs in which both entities appear in safe and unsafe configurations, and evaluation protocols that measure the co-occurrence increment directly. The decomposition introduced here, comparing the score change produced by adding an entity with that produced by moving it, provides a model-agnostic test that can be applied to other relation-defined hazards, such as workers under suspended loads or near unprotected edges.

# 6. Conclusions

This study asked why VLMs misjudge whether a worker is exposed to an excavator struck-by hazard. Using internal probing and interventions on two open-weight VLMs, it separated perception, relation and decision explanations. The evidence answers the four research questions as follows. (RQ1) Both models recognize workers and excavators and encode the near/far relation: object presence is almost perfectly decodable, and the relation is decodable out of fold from late language layers (AUROC 0.82 for Qwen2.5-VL and 0.91 for InternVL3.5). (RQ2) The answers use the relation only weakly: the score moves in the correct direction for most pairs, but far images are answered HAZARD in 74% (Qwen) and 100% (InternVL) of cases, because the score starts from an offset comparable to (Qwen) or nearly eight times (InternVL) the relational shift. (RQ3) The offset is attached to the co-occurrence of the two entities: adding a worker anywhere in an excavator scene raises the hazard score 2.4 and 3.2 times more than moving that worker into the operating zone, and in InternVL the excavator together with the hazard prompt is already sufficient for a HAZARD answer. The same false-alarm pattern appears in natural Rule-4 images. (RQ4) No attention-head group, neuron or direction was found whose removal selectively clears far images, but conditioning a correction on the model's own relation readout improved Qwen's sensitivity–specificity trade-off beyond threshold recalibration.

These findings carry a clear implication for remedies. Because the entities are perceived and the relation is encoded, improving object detection or adding spatial capacity is unlikely to reduce false alarms. Effective remedies must instead strengthen the influence of the encoded relation on the decision or weaken the co-occurrence offset, for example by conditioning the answer on an explicit relation readout or geometric module, by fine-tuning on counterfactual pairs in which the worker and the machine co-occur in both safe and unsafe configurations, and by evaluating models with co-occurring safe images and the co-occurrence increment rather than accuracy alone.

Two limitations bound these conclusions and indicate directions for further work. First, the study examined struck-by exposure through controlled spatial relations in static images. This design isolates the effects of co-occurrence and relation, but real exposure also depends on equipment motion, worker behavior, three-dimensional geometry and operational context. Multi-distance counterfactual series and video are needed to establish physical decision boundaries and to test the account more widely. Second, the study is a first step toward a mechanistic understanding of multimodal reasoning in construction safety, complementing the performance-based evaluations that dominate current research. The same combination of controlled inputs, internal readouts and interventions can be extended to other hazard types and site contexts to examine how visual evidence, linguistic expectation and learned decision biases interact in VLMs, and to develop decision mechanisms that use the spatial evidence these models already encode.

## Declaration of competing interest

None.

## Declaration of generative AI use

The study used OpenAI Codex and Anthropic Claude for coding and writing assistance; the author reviewed and takes responsibility for all content.

## Data availability

Controlled pair manifests, per-image predictions, probe outputs, intervention tables, analysis and plotting scripts will be made available from the corresponding author upon reasonable request.

## Acknowledgements

The author acknowledges the use of the Apollo GPU cluster, a service provided by the Sydney Informatics Hub, a Core Research Facility of the University of Sydney. The author also thanks the Faculty of Engineering, University of Sydney, for start-up funding.